\documentclass[10pt,journal,compsoc]{IEEEtran}
\ifCLASSOPTIONcompsoc
  \usepackage[nocompress]{cite}
\else
  \usepackage{cite}
\fi
\ifCLASSINFOpdf
\else
\fi
\usepackage{amsmath,amsfonts}
\usepackage{array}
\usepackage{textcomp}
\usepackage{stfloats}
\usepackage{url}
\usepackage{verbatim}
\usepackage{graphicx}
\usepackage{multirow}
\usepackage{graphicx}
\usepackage{subfloat}
\usepackage{subfigure}
\usepackage[mathscr]{eucal}
\usepackage{multicol}
\usepackage{algorithm}
\usepackage{algorithmicx}
\usepackage{algpseudocode}
\usepackage{pifont}
\usepackage{float}
\usepackage{multirow}
\usepackage{booktabs}
\usepackage{xcolor,colortbl}
\usepackage{bm}
 \usepackage{utfsym}

\definecolor{gray}{gray}{0.5}

\usepackage{cite}
\usepackage[colorlinks,
            linkcolor=black,
            anchorcolor=black,
            urlcolor=red,
            citecolor=black]{hyperref}

\begin{document}
%
\title{Diffusion-Driven Self-Supervised Learning \\for Shape Reconstruction and Pose Estimation}
%
%
%
%

\author{Jingtao~Sun,
        Yaonan~Wang,
        Mingtao~Feng,
        Chao~Ding,
        Mike Zheng Shou
        and Ajmal Saeed Mian
\IEEEcompsocitemizethanks{\IEEEcompsocthanksitem Jingtao Sun, Yaonan Wang, Mingtao Feng and Chao Ding are with the College of Electrical and Information Engineering and the National Engineering Research Centre for Robot Visual Perception and Control, Hunan University, Changsha, China, 410082.\protect\\
\IEEEcompsocthanksitem Mike Zheng Shou is with Department of Electrical and Computer Engineering and Show Lab, National University of Singapore, Singapore, 117583. Ajmal Mian is with Computer Science and Software Engineering, The University of Western Australia (UWA), Australia, 2006.}
\thanks{Manuscript received December xx, xxxx; revised December xx, xxxx.}
}

%
%

\markboth{Journal of \LaTeX\ Class Files,~Vol.~14, No.~8, August~2015}%
{Shell \MakeLowercase{\textit{et al.}}: Bare Demo of IEEEtran.cls for Computer Society Journals}

\IEEEtitleabstractindextext{%
\begin{abstract}
Fully-supervised category-level pose estimation aims to determine the 6-DoF poses of unseen instances from known categories, requiring expensive mannual labeling costs. Recently, various self-supervised category-level pose estimation methods have been proposed to reduce the requirement of the annotated datasets. However, most methods rely on synthetic data or 3D CAD model for self-supervised training, and they are typically limited to addressing single-object pose problems without considering multi-objective tasks or shape reconstruction. To overcome these challenges and limitations, we introduce a diffusion-driven self-supervised network for multi-object shape reconstruction and categorical pose estimation, only leveraging the shape priors. Specifically, to capture the SE(3)-equivariant pose features and 3D scale-invariant shape information, we present a Prior-Aware Pyramid 3D Point Transformer in our network. This module adopts a point convolutional layer with radial-kernels for pose-aware learning and a 3D scale-invariant graph convolution layer for object-level shape representation, respectively.
Furthermore, we introduce a pretrain-to-refine self-supervised training paradigm to train our network. It enables proposed network to capture the associations between shape priors and observations, addressing the challenge of intra-class shape variations by utilising the diffusion mechanism. Extensive experiments conducted on four public datasets and a self-built dataset demonstrate that our method significantly outperforms state-of-the-art self-supervised category-level baselines and even surpasses some fully-supervised instance-level and category-level methods. The project page is released at \href{https://github.com/S-JingTao/Self-SRPE.git}{Self-SRPE}.
\end{abstract}

}

\maketitle

\IEEEdisplaynontitleabstractindextext

%
\IEEEpeerreviewmaketitle

\IEEEraisesectionheading{\section{Introduction}\label{sec:introduction}}
\IEEEPARstart{O}{bject} 6-DoF (Degrees-of-Freedom) pose estimation is a long-standing problem in computer vision and real-world applications, such as robotics, autonomous driving and virtual reality. Currently, most existing works focus on instance-level methods, which rely heavily on expensive mannual pose labels and precise 3D CAD models. However, collecting ground-truth pose labels and models is typically time-consuming and labor-intensive. In an effort to alleviate the need for extensive labeling, recent research has explored self-supervised~\cite{wang2020self6d,wang2021occlusion,chen2023texpose,hai2023pseudo,gu2022ossid}, semi-supervised~\cite{zhou2021semi} and weakly-supervised~\cite{li2023nerf} methods for pose estimation. Nevertheless, many of these approaches still rely heavily on 3D models and perform suboptimally compared to fully-supervised methods due to the lack of manual pose label.

\begin{figure*}[ht]
    \centering
    \includegraphics[width=\textwidth]{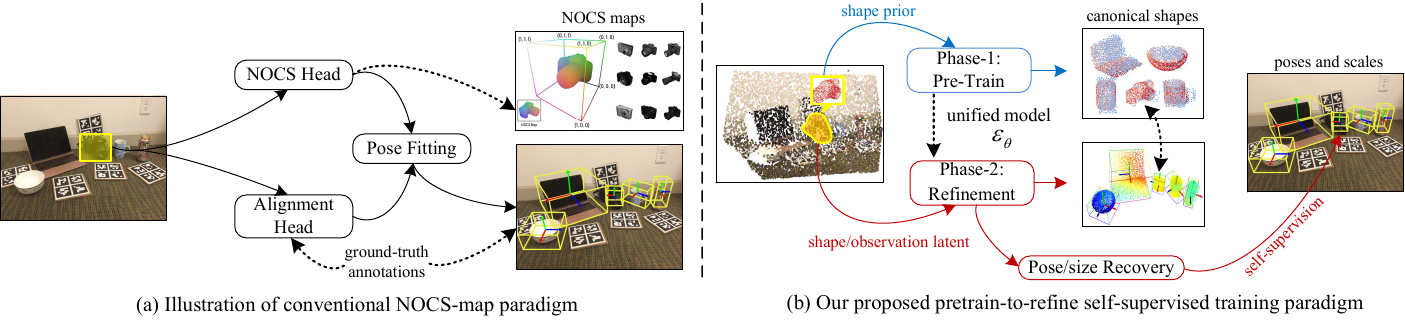} 
    \vspace{-0.4cm}  
    \caption{\textbf{Overview}. Different from most existing NOCS-map paradigm, that employs the one-shot pipeline to normalize objects into a 3D NOCS space contained within a unit cube and aligns their centers or orientations within the same category, our proposed pretrain-to-refine paradigm adopts a two-phase strategy. We first train a basic network model using the prior shapes, and then fine-tune this pre-trained base model under the guidance of the shape/observation latent representations. The entire process is implemented in a self-supervised manner driven by the diffusion mechanism.}\label{FIG_compare}
\end{figure*}

To reduce the dependency on object's 3D CAD model, recent researches have shifted towards the category-level pose estimation. The goal of this task is to estimate size and pose for unseen objects within the same category and the intra-class variations between object shape and texture makes this type of problem more challenging. The pioneering work is NOCS~\cite{wang2019normalized}, as depicted in Fig.~\ref{FIG_compare} (a), which utilizes implicit coordinate space to map all instances into the unified representation.
Despite achieving good results, existing category-level methods such as~\cite{chen2021sgpa,tian2020shape,wang2023query6dof}, typically require fully-supervised training with detailed 6-DoF pose annotations. To address this limitation, recent works~\cite{lee2022uda,lin2022category,peng2022self,liu2023self,li2021leveraging,zaccaria2023self,zhang2022self} have leveraged the self-supervision training with unlabeled synthetic and real-world data for categorical pose estimation.
In~\cite{lee2022uda,lin2022category}, the synthetic data was used to pre-train a deformation network or transferred to real domain using pseudo labels. In~\cite{peng2022self}, 3D CAD models from synthetic data are employed to learn categorical shape information. However, the synthetic-to-real (Sim2Real) domain gap often renders these methods unfeasible and impractical for real-world applications. Alternative available approaches either utilize equivariant shape analysis~\cite{liu2023self,li2021leveraging} or 2D-3D correspondence~\cite{zaccaria2023self,zhang2022self}. Nevertheless, these approaches require strong canonical meshes during the training stage. Most state-of-the-art methods for self-supervised category-level 6-DoF pose estimation focus solely on single objects, rendering them unsuitable for multi-object pose estimation. Moreover, these methods only tackle the pose estimation problem and neglect the corresponding shape reconstruction. In this paper, we extend conventional single-object pose estimation to encompass multi-object shape reconstrcution and categorical pose estimation tasks under a self-supervised manner. Our proposed tasks aim to estimate 6-DoF poses and 3D shapes of multiple surrounding instances in the  observed scene. The main challenges of our proposed tasks are threefold: (i) addressing the fundamental issue of intra-class shape variations for multiple instances, which remains unresolved; (ii) reconstructing 3D shapes for different instances within the same category; (iii) estimating pose and 3D scale without ground-truth pose labels, 3D object's CAD models, or even the Sim2Real setting.

To tackle the above-mentioned challenges, we introduce a diffusion-driven self-supervised network for multi-object shape reconstruction and category-level pose estimation tasks. 
The key challenge of these tasks lies in capturing SE(3)-equivariant pose features and 3D scale-invariant shape information.
To this end, we introduce the Prior-Aware Pyramid 3D Point Transformer module as core structure of our network. Especially, we propose a point convolutional layer with radial-kernels to densely extract category-level SE(3) equivariant pose-aware features for pose estimation, and present a 3D scale-invariant graph convolution layer to learn object-level 3D shape scale-invariant features for shape reconstruction, respectively.
Furthermore, as shown in Fig.~\ref{FIG_compare} (b), to implement self-supervised learning of our proposed network, we introduce a Pretrain-to-Refine Self-Supervised Training paradigm, including two phases. In the pre-training phase, we employ the inverse diffusion mechanism to enable our network to learn the priori knowledge of shape and establish the pre-trained base model. In the refinement phase, we then fine-tune this pre-trained model through self-supervised training by conditioning shape/observation latent embeddings of the object-aware 3D surface information. On top of that, our final refined model captures the shape association between shape priors and current observations to cope with the challenge of intra-class shape variations for every instance.
Experiments on four wildly-used benchmark datasets (REAL275~\cite{wang2019normalized}, CAMERA25~\cite{wang2019normalized}, Wild6D~\cite{ze2022category} and YCB-Video~\cite{xiang2017posecnn}), and our self-built testset DYNAMIC45
demonstrate that our approach can exhibit state-of-the-art performances.
We summarize our contributions as follows:
\begin{itemize}
\item We present a self-supervised network for multi-object shape reconstruction and category-level 6-DoF pose estimation tasks utilizing diffusion mechanism, in which only the shape priors are required.
\item We introduce the Prior-Aware Pyramid 3D Point Transformer module to explore both the SE(3)-equivariant 6-DoF pose features and 3D scale-invariant shape information, effectively capturing the invariant/equivariant property for each object.
\item We propose the Pretrain-to-Refine Self-Supervised Training paradigm, designed to train our network in a self-supervised manner, while effectively solving the challenge of intra-class shape variations.
\item Extensive experiments on four public datasets and our data show that our proposed method achieves state-of-the-art performance for self-supervised category-level pose estimation and comparable results with some category-level fully-supervised approaches. Additionally, our model outperforms some instance-level self/fully supervised baselines.
\end{itemize}
\section{Related Work}
\subsection{Self-Supervised Pose Estimation}
Self-supervised learning has found applications in various domains such as 2D image/video understanding, depth estimation and autonomous driving, etc. Several recent works~\cite{deng2020self,sock2020introducing,wang2020self6d,wang2021occlusion,chen2023texpose,hai2023pseudo,li2023nerf,zhou2021semi,di2021so,gu2022ossid} have investigated self-supervised learning for intance-level pose estimation. In~\cite{deng2020self}, Deng~\emph{et al.} proposed a self-labeling pose estimation mehthod for robotic manipulation. In~\cite{sock2020introducing}, Sock \emph{et al.} leveraged the visual consistency to refine pose estimation under the supervision of ground-truth masks. Self6D~\cite{wang2020self6d} and Self6D++~\cite{wang2021occlusion} were introduced for monocular pose estimation using a synthetic-to-real formulation. 
In contrast to these common render-and-compare piplines, TexPose~\cite{chen2023texpose} decomposed self-supervision for 6D object pose into texture learning and pose learning, while Hai~\emph{et al.}~\cite{hai2023pseudo} proposed a geometry-guided learning framework without any annotations. Some works aim to cover object's pose through weakly-supervised~\cite{li2023nerf} or semi-supervised manners~\cite{zhou2021semi}.
However, these instance-level self-supervised 6D object pose estimation methods can only work with additional depth information or rely on accurate annotation of 2D segmentation masks or object 3D meshes, limiting their application range. 

In contrast, we aim to explore self-supervised category-level pose estimation in this paper. To date, few works have concentrated on self-supervised category-level pose estimation, including~\cite{lee2022uda,peng2022self,zaccaria2023self,lin2022category,liu2023self,zhang2022self,li2021leveraging,he2022towards,manhardt2020cps++,yu2023robotic}. In~\cite{li2021leveraging}, Li~\emph{et al.} proposed a method that leverages SE(3) equivalent representation for category-level pose estimation, while Liu~\emph{et al.}~\cite{liu2023self} presented a part-level SE(3) equivariance for articulated object pose
estimation. CPS++~\cite{manhardt2020cps++} optimized pose results utilizing the consistency between the observed depth map and the rendered depth. Recently, the unsupervised domain adaptation schemes have been used for this task, such as UDA-COPE~\cite{lee2022uda} and Self-DPDN~\cite{lin2022category}. SSC-6D~\cite{peng2022self} leveraged DeepSDF as a 3D representation to improve the pose estimation performance. Additionally, Zaccaria~\emph{et al.}~\cite{zaccaria2023self} utilized 2D optical flow as a proxy for supervising 6D pose.

Our proposed network is also a self-supervised method for category-level 6-DoF pose estimation. However, unlike existing self-supervised category-level approaches that leverage the synthetic data/model, depth image or pseudo labels as additional supervision signals, we only require shape priors to implement self-supervised learning by capturing the association between prior knowledge and observations. This effectively avoids the challenge of sim2real gap. Furthermore, our method not only achieves self-supervised pose estimation but also accomplishes the task of shape reconstruction for multiple objects.

\subsection{Fully-Supervised Pose Estimation}
Previous studies in pose estimation have primarily focused on the problem at the \emph{instance-level}~\cite{shugurov2021dpodv2,zhu2014single,kehl2017ssd,tejani2017latent,wang2021occlusion,peng2019pvnet,guo2021efficient,zhou2020novel,huang2021confidence,wang2021geopose}. 
Few works have specifically addressed \emph{category-level} pose estimation, and the majority of these approaches employ fully supervised learning~\cite{liu2022category,wang2021category,chen2021fs,lin2021dualposenet,liu2022catre,di2022gpv,lin2022sar}. Wang \emph{et al.}~\cite{wang2019normalized} introduced the pioneering work called NOCS, which handles various unseen objects within a category and estimates the 6D pose and dimensions by inferring the correspondence between the observation and the shared NOCS. According to NOCS, Chen \emph{et al.}~\cite{chen2020learning} introduced CASS, a unified representation that models the latent space of canonical shape with normalized pose. Additionally, SPD~\cite{tian2020shape} and SGPA~\cite{chen2021sgpa} employed the shape prior of the observed object to obtain the deformation matrix and structure similarity. Zou \emph{et al.} introduced 6D-ViT in~\cite{zou20226d}, a Transformer-based method that captures the correspondence matrix and deformation field. CenterSnap~\cite{irshad2022centersnap} introduced an approach that simultaneously predicts the 3D shape and estimates the 6D pose and size without relying on bounding boxes. However, all of these methods employ fully-supervised training, relying on ground-truth pose annotations and explicit CAD models during training stage.
Different from these available methods, we address category-level pose estimation problem in a self-supervised manner. Additionally, in contrast to previous works, such as~\cite{chen2021sgpa,tian2020shape,irshad2022centersnap}, we tackle the challenge of intra-class variations by utilizing the diffusion mechanism to capture the shape association between the observations and priors.

\subsection{3D Shape Reconstruction and Completion}
Shape reconstruction and completion present challenges related to generating and estimating objects in 3D space. Initially, researchers attempted to adapt 2D methods to 3D point clouds through techniques such as 3D convolution or voxelization, or by directly processing 3D coordinates. These efforts were inspired by the remarkable performance achieved by PointNet~\cite{qi2017pointnet} and PointNet++~\cite{qi2017pointnet++}. To date, existing approaches can be categorized into point-based (\emph{e.g.,} PCN~\cite{yuan2018pcn}, FinerPCN\cite{chang2021finerpcn}, SK-PCN~\cite{nie2020skeleton}, MSPCN~\cite{zhang2020multi}, and SRPCN~\cite{zhang2022srpcn}), view-based~\cite{hu20203d,gong2021me}, convolution-based, graph-based~\cite{qian2021pu}, transformer-based~\cite{pan20213d,zhao2021point}, and generative model-based methods~(\emph{e.g.,} GAN-based approach~\cite{zhang2021unsupervised} and VAE-based approach~\cite{spurek2022hyperpocket,mittal2022autosdf}). 
What distinguishes our proposed method from these types of 3D shape reconstruction is that we aim to reconstruct the canonical 3D shapes for different instances within the same category, only leveraging the shape priors. While many existing works concentrate on single-instance shape reconstruction with supervision from specific complete point clouds, our method targets multiple objects in current observation and operates in a self-supervised manner without requiring ground-truth point clouds or models.
\begin{table}
\scriptsize
\footnotesize
\setlength{\tabcolsep}{2.0pt}
\centering
\renewcommand\arraystretch{1.0}
\caption{\textbf{Concise notation of symbols.} The main symbols are clarified.}
\label{table_symbol}
\begin{tabular}{l|l}
\toprule [0.8pt]
\multicolumn{1}{c|}{Symbol} & \multicolumn{1}{c}{Description}          \\ \midrule
${P_0} \in {\mathbb{R}^{{N_0} \times 9}}$      & Observable point cloud \\
${P_r} \in {\mathbb{R}^{{N_r} \times 3}}$      & Corresponding point cloud of shape prior \\
$P \in {\mathbb{R}^{K \times M \times 3}}$     & Reconstructed canonical shape point sets \\ 
${{\tilde {\mathcal{P}}_k}} = \{ {{\tilde R}_k},{{\tilde t}_k}\} $  & Estimated 6-DoF poses for all objects \\
${{\varepsilon _\theta }}$ & Learnable parameters of proposed network \\
${{\tilde s}_k} \in {\mathbb{R}^3}$ & Estimated 3D scales for all objects \\
${F_l}$, ${F^{(l)}}(l = 0, 1, 2, 3)$ & Feature map and group map of each resolution \\
$N_x^{(i)}(i = 1, 2, 3, 4)$ & The neighbor domain space\\
${{\tilde x}_k}$ & The coordinates of each radial kenrel points\\
$x_i^t,~t \in (0 \cdots T)$ & Sampled noise points during diffusion process\\
$f$ & The shape/observation latent embedding\\
$|{G_g}|$ & The scale of final pose/size hypothesis\\
\bottomrule [0.8pt]
\end{tabular}
\end{table}

\section{Method}

\begin{figure*}[ht]
    \centering
    \includegraphics[width=\textwidth]{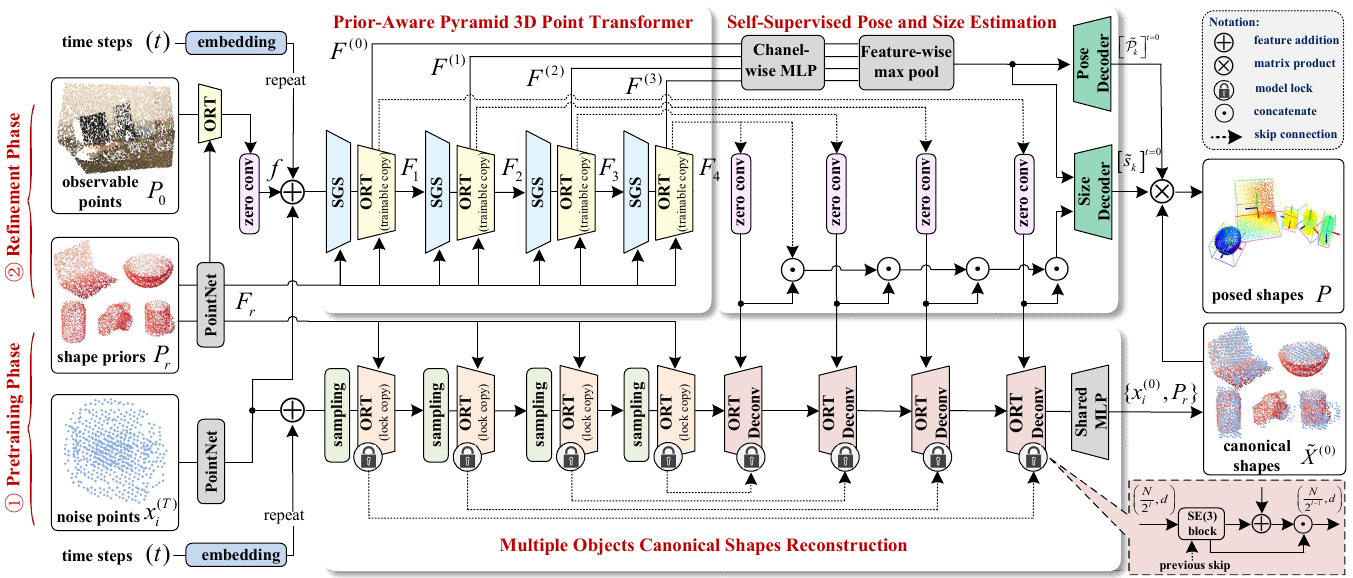}  
    \vspace{-0.3cm} 
    \caption{\textbf{Illustration our network and pretrain-to-refine self-supervised training paradigm}. 
    Our network consists of three components, taking only the observable point cloud ${P_0}$ in the current frame and the corresponding shape priors ${P_r}$ as input. Prior-Aware Pyramid 3D Point Transformer as the core network framework for self-supervised learning.
    During the pre-training phase, we first utilize a simplified version of our Prior-Aware Pyramid 3D Point Transformer with our ORT Deconv to establish a rough base pre-trained model, leveraging the guidance of priori shapes. 
    After that, utilizing the shape/observation latent embedding $f$, we fine-tune this pre-trained model to conduct a comprehensive reinforced model in the subsequent refinement phase. Utimately, the reinforced model is employed to determine the 6-DoF poses, 3D scales and finer canonical shapes.}\label{FIG_overview}
    
\end{figure*}

\subsection{Preliminary}

\textbf{Problem Statement and Notations.}
The goal of this work is to address the simultaneous tasks of self-supervised 3D shape reconstruction and 6-DoF pose localization for all unseen object instances in 3D space.
To clearly demonstrate superior potential of our method, we focus on the stricter self-supervised setting, \emph{w.r.t.,} our model is trained without manual pose annotations, synthetic data conversions and ground-truth CAD object's models. Concretely, given the currently observable partial point cloud ${P_0} \in {\mathbb{R}^{{N_0} \times 9}}$ ($'9'$ contains coordinates, normal vector and color information) and the corresponding shape prior ${P_r} \in {\mathbb{R}^{{N_r} \times 3}}$ for each instance of same category, our objective is to train an end-to-end model ${\Gamma _{{\varepsilon _\theta }}}$ that outputs canonical 3D shape point sets $P \in {\mathbb{R}^{K \times M \times 3}}$ along with the 6-DoF poses ${{\tilde {\mathcal{P}}_k}} = \{ {{\tilde R}_k} \in SE(3),{{\tilde t}_k} \in {\mathbb{R}^3}\} $, and 3D scales ${{\tilde s}_k} \in {\mathbb{R}^3}$ for all currently visible instances:
\begin{equation}\label{equ1}
(P,{\tilde {\mathcal P}_k},{\tilde s_k}) = {\Gamma _{{\varepsilon _\theta }}}({P_0},{P_r}),
\end{equation}
where $k = 1,2, \ldots ,K$, ${K}$ is the number of arbitrary objects in the current scene and ${M}$ denotes the number of points in reconstructed shapes. ${\tilde R}_k$ and ${\tilde t}_k$ are the rotation matrix and translation vector, respectively. ${{\varepsilon _\theta }}$ stands for learnable model parameters. To identify unique shape codes for each shape prior, we employ the Shape Auto-Encoder proposed in~\cite{irshad2022centersnap}.
For the reader's convenience, we clarify the main mathematical symbols in our paper, as provided in Table~\ref{table_symbol}. 

\textbf{Network Architecture.}
As described in Fig.~\ref{FIG_overview}, our network comprises three submodules: (1) Prior-Aware Pyramid 3D Point Transformer serving as the base module for latent representation learning between prior points and current observable point cloud (Sec.~\ref{sec-3.2}); (2) Canonical Shapes Reconstruction module for the coarse-to-fine reconstruction of 3D shapes (Sec.~\ref{sec-3.3}); and (3) Self-Supervised Pose and Size Estimation for final estimation of 6-DoF poses and 3D scales (Sec.~\ref{sec-3.4}).
The following subsections will provide the detailed elaboration of each submodule, demonstrating how our pretrain-to-refine self-supervised training is implemented (refer to Sec.~\ref{sec-3.5}).

\begin{figure*}[ht]
    \centering
    \includegraphics[width=\textwidth]{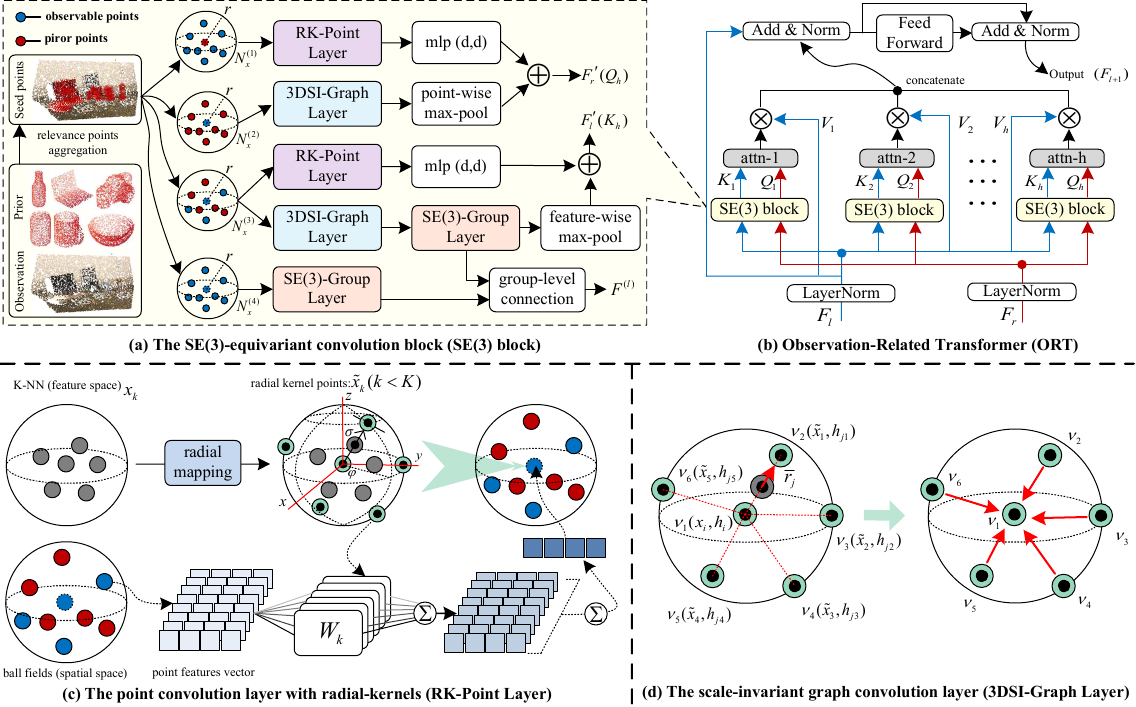} 
    \vspace{-0.3cm}  
      \caption{\textbf{Illustration of the ORT in our proposed Prior-Aware Pyramid 3D Point Transformer.}  \textbf{(a)} The proposed SE(3) block is equipped with an ability to encode shape-observation similarity based on proposed SE(3)-equivariant and 3D scale-invariant learning. \textbf{(b)} The details of our proposed 3D Transformer with multi-head attention, where $h$ is the number of heads. \textbf{(c)} The process of the point convolution layer with radial-kernels, in which the kernel points are generated from the radial mapping on the surface of the sphere domain. \textbf{(d)} The 3D scale-invariant graph convolution layer, that distinguishes itself from other graph convolutions by exhibiting the scale-invariant property through the formation of a graph involving the central point and all radial kernel points.}\label{FIG_ort}
\end{figure*}
\vspace{-0.3cm} 
\subsection{Prior-Aware Pyramid 3D Point Transformer}\label{sec-3.2}
This subsection introduces our proposed Prior-Aware Pyramid 3D Point Transformer, taking ${P_0}$ and ${P_r}$ as inputs for the following discussion. As shown in the top of Fig.~\ref{FIG_overview}, our proposed module includes four transformer encoding stages, generating corresponding feature maps for the observable point clouds at four resolutions. It is noteworthy that, to comprehensively model structural similarities between observable points and shape-prior points, and sufficiently convert semantic features from the prior, we have designed the same structure for all four stages, though the parameters are not shared. Given the observable points ${P_0}$ and prior points ${P_r}$, we first feed them into PointNet-based layer to extract feature embeddings ${F_0} \in {\mathbb{R}^{{N_0} \times {d}}}$ and ${F_r} \in {\mathbb{R}^{{N_r} \times {d}}}$, respectively. After that, these embeddings enter the Transformer ${E_1}$ to produce the feature map ${F_1}$ and group feature map ${F^{(1)}}$. This process is iteratively applied through the subsequent stages ${E_{l}}$ and ${E_{l+1}}$, generating a series of feature maps $\{ {F_1},{F_2},{F_3},{F_4}\} $ with size of ${{({N_0}} \mathord{\left/
 {\vphantom {{({N_0}} {{2^{l+1}}}}} \right.
 \kern-\nulldelimiterspace} {{2^{l+1}}}}) \times d$ and corresponding group feature maps ${F^{(l)}} \in {\mathbb{R}^{({{{N_0}} \mathord{\left/
 {\vphantom {{{N_0}} {{2^{l+1}}}}} \right.
 \kern-\nulldelimiterspace} {{2^{l+1}}}}) \times d \times \left| {{G_g}} \right|}}$ at the four resolutions~($l = 0,1,2,3$). Each transformer stage~${E_{l+1}}$ comprises two blocks: Shape-Guided Sampling (SGS) and Observation-Related Transformer (ORT).

\textbf{Shape-Guided Sampling (SGS).} 
Different from the conventional sampling approach in~\cite{zou20226d,you2022cppf,di2022gpv}, we propose a shape-guided sampling technique to preserve more sampled points associated with the targeted instance, considering both spatial feature distance and semantic similarity between the observed point cloud ${P_0}$ and its shape prior ${P_r}$. Our key insight consists of two aspects: the region of interest of the object in the observation should exhibit a closer feature distance to prior points, and relying solely on the feature distance as the criterion may result in the majority of sampling points falling on the same instance, thus the feature semantic similarity is a nice way to avoid this matter.
Specifically, given the each stage feature map ${F_{l+1}}$ generated from the previous transformer module ${E_{l+1}}$ and the prior point cloud feature map ${F_r}$, we initially calculate the pairwise feature distance mapping matrix $D \in {\mathbb{R}^{({{{N_0}} \mathord{\left/
 {\vphantom {{{N_0}} {{2^{l+1}}}}} \right.
 \kern-\nulldelimiterspace} {{2^{l+1}}}}) \times {N_r}}}$ as follows:
\begin{equation}\label{equ1}
{D_{ij}} = ||{f_i} - f_j^{(r)}|{|_2},\;\forall {f_i} \in {F_{l+1}},\forall f_j^{(r)} \in {F_r},
\end{equation}
where ${\left\|  *  \right\|_2}$ denotes L2-Norm. Subsequently, we compute each distance by considering the overall distance between each point from the current observable point cloud and all prior points in the spatial feature space:
\begin{equation}\label{equ2}
{\nu _i} = \sum\limits_{j = 1}^{{N_r}} {({D_{ij}}),~\forall i \in \{ 1,2, \ldots ,{{{N_0}} \mathord{\left/
 {\vphantom {{{N_0}} {{2^{l+1}}}}} \right.
 \kern-\nulldelimiterspace} {{2^{l+1}}}}\} } .
\end{equation}
We sample a quarter of the points based on the minimum value of ${\nu _i}$, while the remaining quarter sampling points are acquired via the next semantic feature similarity.

The semantic feature similarity is computed using the attention scoring mechanism. We employ the each stage feature map ${F_{l+1}}$ as the group of input vectors $F_{l+1}^i,(i = 1,2, \ldots ,{{{N_0}} \mathord{\left/
 {\vphantom {{{N_0}} {{2^{l+1}}}}} \right.
 \kern-\nulldelimiterspace} {{2^{l+1}}}})$ and take the shape prior feature ${F_r}$ as the query vectors, respectively. The similarity score $\alpha  = [{\alpha _1},{\alpha _2}, \ldots ,{\alpha _{{{{N_0}} \mathord{\left/
 {\vphantom {{{N_0}} {{2^{l+1}}}}} \right.
 \kern-\nulldelimiterspace} {{2^{l+1}}}}}}]$ between $F_{l+1}^i$ and $F_r$ is calculated by the following formula:
\begin{equation}\label{equ3}
\resizebox{.91\hsize}{!}{$
{\alpha _i} = softmax(s(F_{l+1}^i,{F_r})) = \frac{{exp(s(F_{l+1}^i,{F_r}))}}{{\sum\nolimits_{j = 1}^{{{{N_0}} \mathord{\left/
 {\vphantom {{{N_0}} {{2^{l+1}}}}} \right.
 \kern-\nulldelimiterspace} {{2^{l+1}}}}} {exp(s(F_{l+1}^i,{F_r}))} }},$}
\end{equation}
where $s(F_{l+1}^i,{F_r}) = {(F_{l+1}^i)^T} \cdot {F_r}$ is the score function.
We sample the remaining quarter of points based on the maximum value of ${\alpha _i}$.

\textbf{Observation-Related Transformer (ORT).}  
In this block, we aim to explore SE(3)-equivariant pose features and 3D
scale-invariant shape information. Unlike existing works~\cite{chen2021sgpa,tian2020shape}, where the shape prior is directly used to conduct the deformation field and reconstruct the corresponding shape within the normalized space of visible objects, we leverage structure similarity and feature correlation to guide the generation of a unified seed point cloud. Then, a 3D Transformer is constructed to provide intensive characterisation for each resolution.
In particular, given the observable point cloud~${P_l}$ at current stage and its corresponding feature map~${F_l}$, along with the shape prior point cloud~${P_r}$ and its feature map~${F_r}$, we first produce a coarse but complete seed point cloud with a size of $({N_r} + \frac{{{N_0}}}{{{2^l}}}) \times 3$. This seed point cloud captures both the geometry of observations and the structure of the prior shapes. As depicted in Fig.~\ref{FIG_ort} (a), leveraging the feature maps ${F_r}$ and ${F_l}$, the seed generator computes the feature correlation metrics for each individual prior point within the observable point set. We then select the most relevant K points in observable point set as the reference,  with the spatial center coordinate of all reference points serving as the coordinate of the prior point. The aggregation of each priori point facilitates the generation of seed point cloud.
With the seed point cloud, we can define four neighbor domain spaces~${N_x}$ based on the radius $r$, where the domain of definition is the ball, as shown in Fig.~\ref{FIG_ort} (a). For each observable point, there are three domains, while for each prior point, there is one domain, defined as follows:
\begin{align}
\left\{ {\begin{array}{l}
{N_x^{(1)} = \{ {x_i} \in {P_l},x \in {P_r}\left|~ {\left\| {x - {x_i}} \right\|} \right. \le r\} }\\
{N_x^{(2)} = \{ {x_i} \in {P_r},x \in {P_l}\left|~ {\left\| {x - {x_i}} \right\|} \right. \le r\} }\\
{N_x^{(3)} = \{ {x_i} \in {P_r} \cup {P_l},x \in {P_l}\left|~ {\left\| {x - {x_i}} \right\|} \right. \le r\} }\\
{N_x^{(4)} = \{ {x_i},x \in {P_l}\left|~ {\left\| {x - {x_i}} \right\|} \right. \le r\} }
\end{array}} \right..
\end{align}

According to these domains, our network learns the SE(3)-equivariant and 3D scale-invariant properties between observable points and prior points by using our proposed SE(3)-equivariant convolution block (SE(3) block). As depicted in Fig.~\ref{FIG_ort} (a), our SE(3) block takes~${P_l}$, ${P_r}$, ${F_l}$ and ${F_r}$ as inputs, yielding per-point SE(3)-equivariant feature maps ${F_l}^\prime $ and ${F_r}^\prime $, as well as the corresponding group feature map for each level ${F^{(l)}}$. The SE(3) block is composed of three key layers: 1) the point convolution layer with the radial kernels, aggregates point spatial correlation to characterise translation-equivariant properties; 2) the 3D scale-invariant graph convolution layer, captures 3D shape scale-invariance properties; 3) the SE(3) group convolution layer, computes rotation-equivariant correlation between input points and a kernel defined on the continuous SO(3) rotation group~$\{ {G_g}\} $.

\textbf{1) The Point Convolution Layer with Radial-Kernels.} Inspired by the principles of separable convolution in~\cite{chen2021equivariant} and point convolution operation~in~\cite{thomas2019kpconv}, the formulation of this layer can be represented as:
\begin{align}
(\Phi *{h_p})(x,g) = \sum\limits_{{x_i} \in {N_x}} {\Phi ({g^{ - 1}}{x_i},g){h_p}(x - {x_i})} ,
\end{align}
where ${N_x} \in \{ N_x^{(i)},i = 1,2,3,4\}$ and ${h_p}$ is the kernel function under arbitary rotation $g$ defined in the neighbor domain space. 
Due to the kernel is defined in the mixed neighbor space $N_x^{(i)}$, rigid kernel points arranged regularly in~\cite{thomas2019kpconv} are not efficient in this situation. In this regard, we want ${h_p}$ to dynamically adapt to different domain spaces.
As shown in Fig~\ref{FIG_ort} (c), we design a radial mapping operation to generate adaptive kernel points. Our proposed method takes similarity points set of the current point ${x}$ in feature space using KNN as the initial kernel points $\{ {x_k}\left| {k < K} \right.\}  \in {N_x}$, and outputs a set of K radial kernel points ${{\tilde x}_k}(k < K)$ for each initial kernel. The kernel function ${h_p}$ is defined as:
\begin{align}
{h_p}(x - {x_i}) = \sum\limits_{k < K} {\mu (x - {x_i},{{\tilde x}_k})} {W_k},
\end{align}
where  $\{ {W_k}\left| {k < K} \right.\} $ are the associated weight matrices that map features from input size dimension to the output size. Similar to~\cite{thomas2019kpconv}, we use the same linear correlation function, \emph{w.r.t., ${\mu (x - {x_i},{{\tilde x}_k})}$}. For radial kenrel point generation, we project each initial kernel along the radius of ${N_x}$ to map its corresponding radial kernel in the surface of ball field, as shown in Fig.~\ref{FIG_ort}~(c). It is noteworthy that we fix one of the kernel points at the center of the ball. These adaptive radial kernel points not only ensure equivalence to the translation, but also adapt to changes in 3D space of different domains. The coordinates of each radial kenrel ${{\tilde x}_k}$ can be obtained from the following:
\begin{align}
\left\{ {\begin{array}{l}
{{{\tilde x}_k}[x] = r \cdot sin(\sigma ) \cdot cos(\varphi )}\\
{{{\tilde x}_k}[y] = r \cdot sin(\sigma ) \cdot sin(\varphi )}\\
{{{\tilde x}_k}[z] = r \cdot cos(\sigma )}
\end{array}} \right.,
\end{align}
as depicted in Fig.~\ref{FIG_ort}(c), $\sigma$ and $\varphi$ are the angles of latitude and longitude of the radial kernel on the surface of ball field. 

\textbf{2) The 3D Scale-Invariant Graph Convolution Layer.} To preserve the 3D scale-invariant properties of this layer, we first construct a graph $\ell  = (V,\partial )$ with the current point ${x_i}$ and all radial kernel points ${{{\tilde x}_k}}$ as nodes ${\nu _i} \in V$, and edges information $\partial  = ({\alpha _{ij}})$, as depicted in Fig.~\ref{FIG_ort} (d). Here, ${x_i}$ serves as the central point in our graph convolution, and ${\rm N}(i) = \{ j|(i,j) \in \partial \} $ denotes a set of neighbouring points (\emph{radial points}). For each feature node embedding ${h_i},{h_j} \in {\mathbb{R}^{d + 3}}$, we consider both d-dimensional embeddings and its 3D coordinates. To obtain the feature node embeddings ${h_j}$ corresponding to the radial kernel poins, we map the feature embedding of the initial kernel points along the unit radial direction:
\begin{align}
\left\{ {\begin{array}{l}
{{h_j} = linear(h_j^k \otimes {{\bar r}_j})}\\
{{{\bar r}_j} = {{({x_k} - {{\tilde x}_k})} \mathord{\left/
 {\vphantom {{({x_k} - {{\tilde x}_k})} {||{x_k} - {{\tilde x}_k}||}}} \right.
 \kern-\nulldelimiterspace} {||{x_k} - {{\tilde x}_k}||}}}
\end{array}} \right.,
\end{align}
then, we define our scale-invariant graph layer as follows:
\begin{align}\label{h_gnl}
{h_{gnl}} = {\phi _h}({h_i},\sum\limits_{j \in {\rm N}(i)} {{\phi _e}({h_i},{h_j},{\alpha _{ij}})} ),
\end{align}
where ${\phi _h}$ and ${{\phi _e}}$ denote the edge and node operations, respectively, which are commonly approximated by multilayer perceptrons. $h_j^k$ represents the d-dimensional embeddings of initial kernels. Additionally, the mapping of feature node embeddings in Eq.~(\ref{h_gnl}) is only calculated in the unit radial direction and is not affected by spatial distance. Therefore, the 3D scale-invariant property can be directly observed in our proposed layer.

\textbf{3) The SE(3) Group Convolution Layer (SE(3)-Group Layer).} As shown in Fig.~\ref{FIG_ort} (a), the composition of this layer is consistent with~\cite{chen2021equivariant}, aggregating information from neighboring rotation signals within the group. The icosahedron group is considered as the discrete rotation group.
Ultimately, given the per-point SE(3)-equivariant feature maps ${F_l}^\prime $ and ${F_r}^\prime $, we select to leverage the advantage of transformer network~\cite{vaswani2017attention} to model the high-level similarity between the priori shapes and the observable objects. In details, we take ${F_l}^\prime $, ${F_r}^\prime $ and the current stage feature map of the observable points ${F_l}$ as the \emph{query, key} and \emph{value} for the following multi-head attention:
\begin{align}
{\mathcal{A}^{(h)}} = \sigma (\frac{{{F_r}^\prime {W_Q}^{(h)} \cdot {{({F_l}^\prime {W_K}^{(h)})}^T}}}{{\sqrt {{d}} }}){F_l}{W_V}^{(h)},
\end{align}
where, $\sigma ( * )$ denotes the standard softmax normalization function. ${{W_Q}^{(h)}}$, ${{W_K}^{(h)}}$ and ${{W_V}^{(h)}}$ all belong to $ \in {\mathbb{R}^{{d} \times {d}}}$ and are learnbale projection matrics. As depicted in Fig.~\ref{FIG_ort} (b), the final output of the multi-head attention ${F_l}$ is the concatenation of the $h$ individual self-attention outputs ${\mathcal{A}^{(h)}}$, combined with the LayerNorm, Linear layer, ReLU and residual connections.

\begin{figure*}[t]
    \centering
    \includegraphics[width=\textwidth]{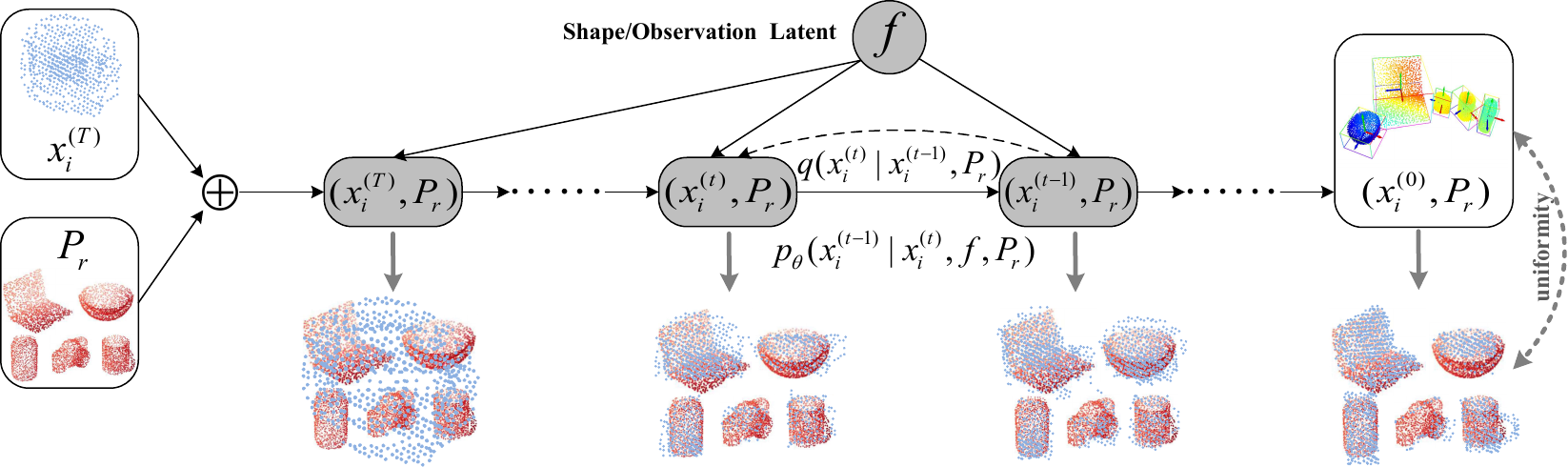} 
    \caption{\textbf{The directed graphical model of diffusion process for canonical shapes reconstruction.} $x_i^{(T)}$ and ${P_r}$ are initial noise points and shape prior points. Each sampled noise points $x_i^{(t)}$ at per timestep contain N points. Related dynamic video can be found in the project page.}\label{FIG_shape}
\end{figure*}
\subsection{Multiple Objects Canonical Shape Reconstruction}\label{sec-3.3}

In this section, the problem of shape reconstruction is treated as the reverse of a conditional diffusion generation process that is modeled as a Markov chain. Given the shape prior ${P_r}$ and the shape/observation latent representation~${f}$, we can reconstruct the canonical shapes from coarse to fine in both two self-supervised training phases.
We start by describing the diffusion formulation, and end with the detailed inference and implementation. For the sake of brevity in the following discussion, we assume that each sampled noise point cloud at per timestep contains a set of $N$ points with xyz-coordinates, denoted as ${X^{(t)}} = \{ \{ x_i^{(t)} \in {\mathbb{R}^3}\} _{i = 1}^N\} _{t = 0}^T$.

\textbf{Diffusion Formulation.} 
As displayed in Fig.~\ref{FIG_shape}, we firstly take the poins from a simple noise distribution $x_i^{(T)} \sim \mathcal N (0,{\rm I})$ and the shape prior point clouds ${P_r}$ as the inputs to construct the initial noise points~${{\tilde X}^{(T)}} = (x_i^{(T)},{P_r})$. These noise points are passed through the reverse Markov chain to form the complete sharp shapes regarded as the ${{\tilde X}^{(0)}} = ( x_i^{(0)},{P_r}) _{i = 1}^N$. Following the conventional diffusion formulations in~\cite{ho2020denoising}, we describe the \emph{ conditional forward diffusion process} with the number of timesteps $(t = 1,2,3, \ldots ,T)$, which gradually adds the noise to $ x_i^{(0)}$ while maintaining ${P_r}$ unchanged. This process yields a series of shape variables with decreasing noise distribution levels, denoted as $\{ (x_i^{(1)},{P_r}),(x_i^{(2)},{P_r}), \ldots ,(x_i^{(T)},{P_r})\} $. To learn our diffusion process, we define the ground-truth diffusion distribution ${q(*)}$ by gradually adding Gaussian noise to the prior's shapes:
\begin{align}\label{eq_forward}
q(x_i^{(1:T)}|x_i^{(0)}) = \prod\limits_{t = 1}^T {q(x_i^{(t)}|x_i^{(t - 1)},{P_r})} ,
\end{align}
where ${q(x_i^{(t)}|x_i^{(t - 1)},{P_r})}$ is the Markov diffusion kernel defined as:
\begin{align}\label{eq_forward}
q(x_i^{(t)}|x_i^{(t - 1)},{P_r}) = \mathcal{N}(x_i^{(t)}|\sqrt {1 - {\beta _t}} x_i^{(t - 1)},{\beta _t}{\rm I}),
\end{align}
with $\beta _t$ representing the variance schedule parameter controlling the step size, $\{ {\beta _t} \in (0,1)\} _{t = 1}^T$.

Since the reverse process aims to recover the expected shapes from the input noise, requiring training from data, we formulate \emph{the reverse process} as a conditional denoising process:
\begin{align}\label{eq_reverse}
{p_\theta }(x_i^{(0:T)}|f,{P_r}) = {{p_\theta }}(x_i^{(T)})\prod\limits_{t = 1}^T {{p_\theta }(x_i^{(t - 1)}|x_i^{(t)},f,{P_r})} ,
\end{align}
\begin{align}\label{reverse}
\resizebox{.89\hsize}{!}{$
{p_\theta }(x_i^{(t - 1)}|x_i^{(t)},f,{P_r}): = \mathcal{N}(x_i^{(t - 1)}|{\mu _\theta }(x_i^{(t)},f,{P_r},t),{\beta _t}{\rm{I}}),$}
\end{align}
where ${\mu _\theta }$ represents the shape prediction parameterized by $\theta$, and ${{p_\theta }(x_i^{(T)})}$ is a standard Gaussion prior. It is worth highlighting that the reverse process maintains the shape prior $P_r$ fixed and diffuses only the deformation part of inter-category shape variation between visible instances in observations and the shape priors with the same category.

\textbf{Inference Details.}
As depicted in the bottom of Fig.\ref{FIG_overview}, to align with our proposed Prior-Aware Pyramid 3D Point Transformer, we adapt a unified pyramid structure for the module design of shape reconstruction task.
Besides, traditional downsampling approach is directly applied without emloying our SGS, owing to the input comprises $x_i^{(t)}$ and ${{P_r}}$. These input points originally constitute the seed point cloud, lacking practical significance in terms of noise point features.
We also add the temporal embeddings into our network and use a sinusoidal position embedding along with a MLP to obtain these temporal embeddings, that contain sine and cosine pairs with varying frequencies:
\begin{align}
\resizebox{.89\hsize}{!}{$
mlp(sin({\varpi _1}t),cos({\varpi _1}t), \ldots ,sin({\varpi _{{d \mathord{\left/
 {\vphantom {d 2}} \right.
 \kern-\nulldelimiterspace} 2}}}t),cos({\varpi _{{d \mathord{\left/
 {\vphantom {d 2}} \right.
 \kern-\nulldelimiterspace} 2}}}t)),$}
\end{align}
where ${\varpi _i} = 1/({10^{{{i} \mathord{\left/
 {\vphantom {{i} d}} \right.
 \kern-\nulldelimiterspace} d}}})$. Meanwhile, $t$ and $d$ indicate the timestep prompt and the feature dimension, respectively.

During the pretrain phase, with the priori shapes ${P_r}$, rough shapes can be reconstructed through four ORT Deconvolution (ORT-Deconv) blocks along with a shared-MLP. In our ORT-Deconv, we use the simple SE(3) block without the branch of group features extraction as the base submodule. Output features from the downsampled point set are mapped to a higher-resolution set via concatenation with the features from skip connections, as displayed in the bottom right of Fig.\ref{FIG_overview}. During the refinement phase, the shape/observation latent representation $f$ is added into the trainable copy of four Transformer block in Prior-Aware Pyramid 3D Point Transformer, using a $1 \times 1$ zero convolution layer initialized with  both weight and bias set to zeros. Their outputs are added to four skip-connectons and then passed through four zero convolution layers, which are subsequently connected with the corresponding ORT-Deconv. The refiner cannoical shapes can be obtained using the final refined model. Detailed inference for shape reconstruction can be found in Alg.~\ref{alg::inference}.

\begin{algorithm}[ht] 
  \caption{Self-Supervised Training} 
  \label{alg::train}
  \begin{algorithmic}[1] 
    \Require
    ${T}$: total timesteps;
    \Require
    $P_r$: corresponding shape prior sets;
    \Require
    $P_0$: original observable point cloud;
    \Require
    $D_{data}(P_0)$: partial point cloud dataset;\newline
    \renewcommand{\algorithmicensure}{\textbf{\ding{172}~Pre-training Phase:}} 
    \Ensure
    \Repeat 
    \State Sample ${X^{(0)}} \sim {D_{data}}(P_0) \to x_i^{(0)}$;
    \State Sample t $ \sim $ Uniform($\{ 1,2, \ldots ,T\} $);
    \State Sample $\varepsilon \sim \mathcal{N}(0,I)$ and ${\beta _t}$;
    \State Sample ${\beta _t}$;
    \State Compute ${{\tilde \alpha }_t} = \prod\nolimits_{s = 1}^t {(1 - {\beta _s})} $;
    \State Compute $x_i^{(t)} = \sqrt {{{\tilde \alpha }_t}} x_i^{(0)} + \sqrt {1 - {{\tilde \alpha }_t}} \varepsilon $;
    \State  Perform gradient descent with Eq.~(\ref{L-pre}): \newline
     \begin{center}
    ${\nabla _\theta }(\sum\nolimits_{i = 1}^N {||\varepsilon  - {\varepsilon _\theta }(x_i^{(t)},{P_r},t)||}) $;
     \end{center} 
    \Until converged\newline
    \renewcommand{\algorithmicensure}{\textbf{\ding{173}~Refinement Phase:}} 
    \Ensure
    \Repeat 
    \State Sample ${X^{(0)}} \sim {D_{data}}(P_0) \to x_i^{(0)}$;
    \State Sample t $ \sim $ Uniform($\{ 1,2, \ldots ,T\} $);
    \State Sample $\varepsilon \sim \mathcal{N}(0,I)$;
    \State Sample $x_i^{(t)}$ from previous phase;
    \State Sample $P_0$;
    \State Learn shape/observation latent $f$ from $P_0$ and $P_r$;
    \State Perform gradient descent and fine-tune with Eq.~(\ref{L-all}): \newline
     \begin{center}
    ${\nabla _\theta }(\sum\nolimits_{i = 1}^N {||\varepsilon  - {\varepsilon _\theta }(x_i^{(t)},f,{P_r},t)||}) $; 
     \end{center}
    \Until converged
  \end{algorithmic} 
\end{algorithm}

\begin{algorithm}[ht] 
  \caption{Inference} 
  \label{alg::inference}
  \floatname{algorithm}{Procedure}
  \renewcommand{\algorithmicrequire}{\textbf{Input:}}
  \renewcommand{\algorithmicensure}{\textbf{Output:}} 
  \begin{algorithmic}[1] 
    \Require
    total timesteps: $T$;
    \Require
    shape prior point: $P_r$;
    \Require
    current observable point cloud: $P_0$;
    \Require
    noise distribution: ${X^{(T)}} \leftarrow \{ x_i^{(T)}\}  \sim \mathcal{N}(0,I)$;
    \Ensure
    6-DoF poses: ${\tilde {\mathcal{P}}_k} = \{ {{\tilde R}_k},{{\tilde t}_k}\} $ and 3D scales: ${{\tilde s}_k}$;
    \Ensure
    posed shapes after reconstruction: $P = \{ {P_k}\} _{k = 1}^K$;\newline
    \renewcommand{\algorithmicrequire}{\textbf{$ * $~Canonical Shapes Reconstruction:}}
    \Require
    \For 
    {$t = T,T - 1, \ldots ,1$}

    \State Sample $z \sim \mathcal{N}(0,I)$ if $t > 1$ else $z = 0$;
    \State Sample ${\beta _t}$ and ${\alpha _t} = 1 - {\beta _t}$;
    \State Compute ${{\tilde \alpha }_t} = \prod\nolimits_{s = 1}^t {(1 - {\beta _s})} $ and ${\sigma _t} = \frac{{1 - {{\tilde \alpha }_{t - 1}}}}{{1 - {{\tilde \alpha }_t}}}{\beta _t}$;
    \State Compute points distribution in each timestep:\newline
    \begin{center}
    $(x_i^{(t - 1)},{P_r}) = \frac{1}{{\sqrt {{\alpha _t}} }}(x_i^{(t)} - \frac{{{\beta _t}}}{{\sqrt {1 - {{\tilde \alpha }_t}} }}{\varepsilon _\theta }(x_i^{(t)},f,{P_r},t)) + {\sigma _t}z$;
    \end{center}
    \EndFor 
    \State Sample ${{\tilde X}^{(0)}} = \{ x_i^{(0)},{P_r}\} $ and ${{\tilde X}^{(0)}} = \{ \tilde X_k^{(0)}\} _{k = 1}^K$;\newline
    \renewcommand{\algorithmicrequire}{\textbf{$ * $~Pose/size Estimation and Shapes Recovery:}}
    \Require

    \State Set timestep $t = 0$;
    \State Recover the assumption subsets $\{ {{\tilde {\mathcal{P}}}^{(i)}}\} $ and $\{ {{\tilde s}^{(j)}}\} $;
    \For 
    {$k = 1,2, \ldots ,K$}
    \For
    {$i = 1,2, \ldots ,|{G_g}|$}
    \For
    {$j = 1,2, \ldots ,|{G_g}|$}
    \State Pose $P_k^{(i,j)}$ with Eq.~(\ref{L-posed});
    \EndFor 
    \EndFor 
    \EndFor 
    \State Calculate the distance $d(P_k^{(i,j)},P_0^k)$ with Eq.~(\ref{distance});
    \State Select optimum $\tilde {\mathcal{P}}_k^{(i^*)}$, $\tilde s_k^{(j^*)}$ and $P_k^{(i^*,j^*)}$ with Eq.~(\ref{optimum});
  \end{algorithmic}
  \Return ${\tilde {\mathcal{P}}_k}$, ${{\tilde s}_k}$ and $P$.
\end{algorithm}
\subsection{Self-Supervised Pose and Size Estimation}\label{sec-3.4}
Having the final trained model while the timestep $t$ is set to zero, we introduce two self-supervised estimation branches:

\textbf{6-DoF Poses Estimation Branch.}
The objective of this branch is to develop an estimation decoder from the per-group features map ${F^{(l)}}$ to the SE(3)-equivariant 6-DoF poses for each visible objects. Initially, we aim to predict a subset of pose assumptions including every 6-DoF pose hypothesis for corresponding icosahedron group within $\left\{ {{G_g}} \right\}$. Here, this estimation decoder maps ${F^{(l)}} \in {\mathbb{R}^{({{{N_0}} \mathord{\left/
 {\vphantom {{{N_0}} {{2^{l+1}}}}} \right.
 \kern-\nulldelimiterspace} {{2^{l+1}}}}) \times d \times \left| {{G_g}} \right|}}$ to $\{ {{\tilde {\mathcal{P}}}^{(i)}}\}  \in {\mathbb{R}^{K \times (4 + 3) \times |{G_g}|}},(i = 1,2, \ldots ,|{G_g}|)$, and here $4$ represents the rotation vector $\tilde R_k^{(i)}$ acted by a quaterion $\tilde q_k^{(i)}$ and $3$ denotes the translation vector $\tilde t_k^{(i)}$. To implement this goal, a channel-wise MLP is applied: $f \in {\mathbb{R}^{{N_0}/{2^{l+1}}}} \to f \in {\mathbb{R}^K}$, operating solely on each group features map ${F^{(l)}}$ individually, yielding a new set of features. Then, the feature-wise max-pooling over $\{ {F^{(l)}}\} $ is performed, followed by an MLP-based pose decoder to obtain the final pose assumption subset $\{ {{\tilde {\mathcal{P}}}^{(i)}}\} $.

\textbf{3D Sizes Recovery Branch.} The objective of this branch is to learn an SE(3)-invariant 3D sizes recovery decoder for every observable objects, capable of predicting the per-object's size from the feature map $F_4$. Initially, the recovery function maps $F_4$ to another size assumptions subset $\{ {{\tilde s}^{(j)}}\}  \in {\mathbb{R}^{K \times 1 \times |{G_g}|}},(j = 1,2, \ldots ,|{G_g}|)$ containing per-object per-scale for corresponding icosahedron group within $\left\{ {{G_g}} \right\}$, where $1$ denotes the value for per-object 3D scale $\tilde s_k^{(j)}$. To achieve 3D scale-invariant, both feature map $F_4$ and the corresponding feature after zero convolution are concatenated, as shown in the top right of Fig.~\ref{FIG_overview}. 
This concatenated output undergoes similar operations in the next three stages, along with an MLP, to obtain a base feature map with the size of $K \times d$. Subsequently, this process is repeated and added to the generated features from the pose estimation branch, and an MLP-based size decoder is used to obtain the final size assumption subset ${ {{\tilde s}^{(j)}}} $.

Given both pose and size assumption subsets, we can pose the canonical shapes point cloud using each assumption $\tilde {\mathcal{P}}_k^{(i)}$ and $\tilde s_k^{(j)}$, rendering posed reconsrtuction shape points for per-object and every hypotheses:
\begin{align}\label{L-posed}
P_k^{(i,j)} = \tilde s_k^{(j)} \cdot \tilde R_k^{(i)} \cdot \tilde X_k^{(0)} \cdot {(\tilde R_k^{(i)})^{ - 1}} + \tilde t_k^{(i)}.
\end{align}
This process involves calculating the point cloud distance between them and the original input of partial observation ${P_0} = \{ P_0^k\} _{k = 1}^K$:
\begin{equation}\label{distance}
\begin{aligned}
d(P_k^{(i,j)},P_0^k) 
&= \sum\limits_{k = 1}^K {(\sum\limits_{x \in P_k^{(i,j)}} {\mathop {\min }\limits_{y \in P_0^k} } ||x - y|{|^2}} \\
&+ {\sum\limits_{y \in P_0^k} {\mathop {\min }\limits_{x \in P_k^{(i,j)}} } ||x - y|{|^2})},
\end{aligned}
\end{equation}
we select the best pose and size assumption values, and correponding posed shape points as the final results, that satisfy the minimum distance:
\begin{align}\label{optimum}
({i^*},{j^*}) = \mathop {\arg \min }\limits_{i,j \in \{ 1,2, \ldots ,|{G_g}|\} } d(P_k^{(i,j)},P_0^k).
\end{align}
Further details on inference process for pose and size can be found in the last part of Alg.~\ref{alg::inference}.

\subsection{Pretrain-to-Refine Self-Supervised Training}\label{sec-3.5}
To enable the self-supervised training of our proposed network, we introduce a \emph{Pretrain-to-Refine Self-Supervised Training} paradigm depicted in Fig.~\ref{FIG_overview}, inspired by ControlNet~\cite{zhang2023adding}. At the pre-training phase, we consider the canonical 3D shapes reconstruction task as a conditional diffusion generation problem where the shape priors~${P_r}$ serve as the basic conditioner. 
We build a base model with the similar network structure of Prior-Aware Pyramid 3D Point Transformer (the bottom left of Fig.~\ref{FIG_overview}). At the reinforcement phase, we lock all parameters ${{\varepsilon _\theta }}$ of this model and clone it into a trainable copy, namely, per-block of Prior-Aware Pyramid 3D Point Transformer. We add the observable/shape latent embedding $f$ as a task-specific extra condition to fine-tune the  base model. Further detailed of this process can be found in Alg.~\ref{alg::train}. 

\textbf{Training Objective.}
Having formulated the conditional forward and reverse diffusion processes in Eq.~(\ref{eq_forward}) and (\ref{reverse}), we now formalize the self-supervised training objective as follows. Similarly to DDPM~\cite{ho2020denoising}, we also use the variational lower bound to optimize the negative log-likelihood of the point belonging to desired shape~${\mathbb{E}_q}\log {p_\theta }(x_i^{(0)})$:
\begin{align}\label{TO}
 - {\mathbb{E}_q}\log {p_\theta }(x_i^{(0)}) \le {\mathbb{E}_q}\left[ {\log \frac{{q(x_i^{(1:T)}|x_i^{(0)})}}{{{p_\theta }(x_i^{(0:T)}|f,{P_r})}}} \right] = {L},
\end{align}
according to the process in~\cite{sohl2015deep} and considering the set of all shape points containing ${N}$ points, the above objective can be further adapted to be a series of KL-divergence and entropy terms and the detailed process can be found in Appendix~\ref{appendix-a}:
\begin{align}\label{adapt-main}
L = {\mathbb{E}_q}\left[ {\begin{array}{*{20}{l}}
{\sum\limits_{t = 2}^T {\sum\limits_{i = 1}^N {\underbrace {\left( {\begin{array}{*{20}{l}}
{{D_{KL}}(q(x_i^{(t - 1)}|x_i^{(t)},x_i^{(0)})||}\\
{{p_\theta }(x_i^{(t - 1)}|x_i^{(t)},f,{P_r}))}
\end{array}} \right)}_{{L_{t - 1}}}} } }\\
{ + \underbrace {{D_{KL}}(q({X^{(T)}}|{X^{(0)}})||{p_\theta }({X^{(T)}}))}_{{L_T}}}\\
{ - \sum\limits_{i = 1}^N {\underbrace {\log {p_\theta }(x_i^{(0)}|x_i^{(1)},f,{P_r})}_{{L_0}}} }
\end{array}} \right].
\end{align}
Finally, in order to be consistent with the simplified option proposed in~\cite{ho2020denoising}, we reparameterize our final refinement objective function by reducing Eq.~(\ref{adapt-main}) to an $L_{2}$ loss:
\begin{align}\label{L-all}
{L_{refine}} = \mathbb{E}\left[ {\sum\nolimits_{i = 1}^N {||\varepsilon  - {\varepsilon _\theta }(x_i^{(t)},f,{P_r},t)||} } \right].
\end{align}
Similarly, our pre-training objective function can be simply modified to learn a conditional generative model given prior shapes. We modifies this generative model by conditioning the step function ${\varepsilon _\theta }$ on the input tensor that combines information derived from both the current state $x_i^{(t)}$ and the shape prior ${{P_r}}$:
\begin{align}\label{L-pre}
{L_{pretrain}} = \mathbb{E}\left[ {\sum\nolimits_{i = 1}^N {||\varepsilon  - {\varepsilon _\theta }(x_i^{(t)},{P_r},t)||} } \right].
\end{align}

\section{Experiments}

In this section, we conduct extensive experiments to evaluate the performance of our method and compare it with currently available state-of-the-art baselines on five datasets, including four public datasets and our self-built dataset.

\begin{table*}[ht]
\scriptsize
\footnotesize
\setlength{\tabcolsep}{1.5pt}
\centering
\renewcommand\arraystretch{1.0}
\caption{\textbf{Quantitative comparison of fully-supervised category-level pose estimation on the pubilc CAMERA25 and REAL275 dataset.} Note that the best and the second best results are highlighted in \textbf{bold} and \underline{underlined}. The comparison results of current state-of-the-art baselines are summarized from their original papers. Empty entries indicate that they were not reported in the original paper.}
\label{table_full}

\begin{tabular*}{\textwidth}{@{\extracolsep{\fill}}cl|cccccc|cccccc}
\toprule [0.8pt] 
\multicolumn{2}{c|}{}                                                                                & \multicolumn{6}{c|}{CAMERA25 Dataset}                                                                                                                                    & \multicolumn{6}{c}{REAL275 Dataset}                                                                                                                                      \\  
\multicolumn{2}{c|}{\multirow{-2}{*}{Method}}                                                        & $IoU50$                  & \multicolumn{1}{c|}{$IoU75$}                  & ${5^ \circ }2{\rm{cm}}$                  & ${5^ \circ }5{\rm{cm}}$                  & ${10^ \circ }2{\rm{cm}}$                 & ${10^ \circ }5{\rm{cm}}$                 & $IoU50$                  & \multicolumn{1}{c|}{$IoU75$}                  & ${5^ \circ }2{\rm{cm}}$                  & ${5^ \circ }5{\rm{cm}}$                  & ${10^ \circ }2{\rm{cm}}$                 & ${10^ \circ }5{\rm{cm}}$                 \\ \midrule
1                                       &  NOCS~\cite{wang2019normalized}~{\color{gray}[CVPR2019]}                  & 83.9                   & \multicolumn{1}{c|}{69.5}                   & 32.3                   & 40.9                   & 48.2                   & 64.6                   & 80.5                   & \multicolumn{1}{c|}{30.1}                   & 7.2                    & 10.0                   & 13.8                   & 25.2                   \\
2                                       & SDP~\cite{tian2020shape}~{\color{gray}[ECCV2020]}                                           & 93.2                   & \multicolumn{1}{c|}{83.1}                   & 54.3                   & 59.0                   & 73.3                   & 81.5                   & 77.3                   & \multicolumn{1}{c|}{53.2}                   & 19.3                   & 21.4                   & 43.2                   & 54.1                   \\
3                                       & FS-Net~\cite{chen2021fs}~{\color{gray}[CVPR2021]}                                       & -                      & \multicolumn{1}{c|}{85.2}                   & -                      & 62.0                   & -                      & 60.8                   & 81.1                   & \multicolumn{1}{c|}{52.0}                   & 19.9                   & 33.9                   & -                      & 69.1                   \\
4                                       & SGPA~\cite{chen2021sgpa}~{\color{gray}[ICCV2021]}                                         & 93.2                   & \multicolumn{1}{c|}{88.1}                   & 70.7                   & 74.5                   & 82.7                   & 88.4                   & 80.1                   & \multicolumn{1}{c|}{61.9}                   & 35.9                   & 39.6                   & 61.3                   & 70.7                   \\
5                                       & CenterSnap~\cite{irshad2022centersnap}~{\color{gray}[ICRA2022]}                                   & 92.5                   & \multicolumn{1}{c|}{-}                      & -                      & 66.2                   & -                      & 81.3                   & 80.2                   & \multicolumn{1}{c|}{-}                      & -                      & 27.2                   & -                      & 58.8                   \\
6                                       & ShAPO~\cite{Irshad2022ShAPOIR}~{\color{gray}[ECCV2022]}                                        & 93.5                   & \multicolumn{1}{c|}{-}                      & -                      & 66.6                   & -                      & 81.9                   & 79.0                   & \multicolumn{1}{c|}{-}                      & -                      & 48.8                   & -                      & 66.8                   \\
7                   & SAR-Net~\cite{lin2022sar}~{\color{gray}[CVPR2022]}                                      & 86.8                   & \multicolumn{1}{c|}{79.0}                   & 66.7                   & 70.9                   & 75.3                   & 80.3                   & 79.3                   & \multicolumn{1}{c|}{62.4}                   & 31.6                   & 42.3                   & 50.3                   & 68.3                   \\
8                   & GPV-Pose~\cite{di2022gpv}~{\color{gray}[CVPR2022]}                                     & 92.9                   & \multicolumn{1}{c|}{86.6}                   & 67.4                   & 76.2                   & -                      & 87.4                   & \underline{83.0}             & \multicolumn{1}{c|}{64.4}                   & 32.0                   & 42.9                   & -                      & 73.3                   \\
9                   & HS-Pose~\cite{zheng2023hs}~{\color{gray}[CVPR2023]}                                      & 93.3                   & \multicolumn{1}{c|}{\underline{89.4}}             & 73.3                   & 80.5                   & 80.4                   & 89.4                   & 82.1                   & \multicolumn{1}{c|}{74.7}                   & 46.5                   & 55.2                   & 68.6                   & 82.7                   \\
10                  & IST-Net~\cite{liu2023prior}~{\color{gray}[ICCV2023]}                                      & \underline{93.7}             & \multicolumn{1}{c|}{90.8}                   & 71.3                   & 79.9                   & 79.4                   & 89.9                   & 82.5                   & \multicolumn{1}{c|}{\underline{76.6}}             & 47.5                   & 53.4                   & \underline{72.1}             & 80.5                   \\
11                  & Query6DoF~\cite{wang2023query6dof}~{\color{gray}[ICCV2023]}                                    & 91.9                   & \multicolumn{1}{c|}{88.1}                   & \underline{78.0}             & \underline{83.1}             & \underline{83.9}             & \underline{90.0}             & 82.5                   & \multicolumn{1}{c|}{76.1}                   & 46.8                   & 54.7                   & 67.9                   & 81.6                   \\
12                  & VI-Net~\cite{lin2023vi}~{\color{gray}[ICCV2023]}                                       & -                      & \multicolumn{1}{c|}{79.1}                   & 74.1                   & 81.4                   & 79.3                   & 87.3                   & -                      & \multicolumn{1}{c|}{48.3}                   & \underline{50.0}             & \underline{57.6}             & 70.8                   & \underline{82.1}             \\
13                  & GPT-COPE~\cite{zou2023gpt}~{\color{gray}[TCSVT2023]}                                    & 92.5                   & \multicolumn{1}{c|}{86.9}                   & 70.4                   & 76.5                   & 81.3                   & 88.7                   & 82.0                   & \multicolumn{1}{c|}{70.4}                   & 45.9                   & 53.8                   & 63.1                   & 77.7                   \\ 
14 & MH6D~\cite{liu2024mh6d}~{\color{gray}[TNNLS2024]}                                    & -                   & \multicolumn{1}{c|}{-}                   & -                   & -                   & -                   & -                   & 70.2                   & \multicolumn{1}{c|}{45.0}                   & 40.4                   & 50.1                  & 58.3                   & 73.1                   \\ \midrule
\multicolumn{2}{l|}{Ours}                                                                            & \textbf{94.5}          & \multicolumn{1}{c|}{\textbf{92.1}}          & \textbf{79.4}          & \textbf{85.2}          & \textbf{88.5}          & \textbf{92.0}          & \textbf{86.1}          & \multicolumn{1}{c|}{\textbf{77.1}}          & \textbf{55.0}          & \textbf{59.8}          & \textbf{73.7}          & \textbf{83.7}          \\ \bottomrule [0.8pt]
\end{tabular*}
\end{table*}

\begin{table*}[ht]
\scriptsize
\footnotesize
\setlength{\tabcolsep}{4.0pt}
\centering
\renewcommand\arraystretch{1.0}
\caption{\textbf{Quantitative comparison of self-supervised category-level pose estimation on the pubilc REAL275 dataset.} Note that the best and the second best results are highlighted in \textbf{bold} and \underline{underlined}. The comparison results of other baselines are summarized from their original papers, and empty entries indicate that they were not reported. "Syn." and "Real" denote the uses of training data of synthetic CAMERA25 and real-world REAL275 datasets, respevtively. "Multiple" and "Single" indicate whether the method is for the single object or for multiple objects.}
\label{table_self}
\begin{tabular}{l|ccc|cc|c|cccccc}
\toprule [0.8pt] 
\multicolumn{1}{c|}{\multirow{2}{*}{Method}} & \multicolumn{3}{c|}{Training Data} & \multicolumn{2}{c|}{Objective} & \multirow{2}{*}{Shape Prior} & \multicolumn{6}{c}{Evaluation Metrics}                                                        \\
\multicolumn{1}{c|}{}                        & Input      & Real      & Syn.      & Multiple        & Single       &                                                                        & $IoU50$         & $IoU75$         & ${5^ \circ }2{\rm{cm}}$         & ${5^ \circ }5{\rm{cm}}$         & ${10^ \circ }2{\rm{cm}}$         & ${10^ \circ }5{\rm{cm}}$         \\ \midrule
SSC-6D~\cite{peng2022self}~{\color{gray}[AAAI2022]}                        & RGB-D      & \usym{2713}         & \usym{2713}         &                 & \usym{2713}            & \usym{2713}                                                                      & 72.9         & -             & 16.8          & 19.6          & 44.12         & 54.5          \\
Self-DPDN~\cite{lin2022category}~{\color{gray}[ECCV2022]}                     & RGB-D      & \usym{2713}         & \usym{2713}         &                 & \usym{2713}            & \usym{2713}                                                                      & \underline{83.0}    & \underline{70.3}    & \underline{39.4}    & \underline{45.0}    & \underline{63.2}    & \underline{72.1}    \\
UDA-COPE~\cite{lee2022uda}~{\color{gray}[CVPR2022]}                      & RGB-D      & \usym{2713}         & \usym{2713}         &                 & \usym{2713}            & \usym{2717}                                                                      & 82.6          & 62.5          & 30.4          & 34.8          & 56.9          & 66.0          \\
Zaccaria et al.~\cite{zaccaria2023self}~{\color{gray}[RAL2023]}                 & RGB-D      & \usym{2713}         & \usym{2713}         &                 & \usym{2713}            & \usym{2717}                                                                      & 75.2         & -             & 34.9          & 39.5          & 53.1          & 63.8          \\
Self-Pose~\cite{zhang2022self}~{\color{gray}[ICLR2023]}                     & RGB        & \usym{2713}         & \usym{2717}         &                 & \usym{2713}            & \usym{2713}                                                                      & 41.7          & -             & -             & 11.6          & -             & 28.3          \\
SCNet~\cite{yu2023robotic}~{\color{gray}[T-MECH2023]}                       & RGB-D      & \usym{2713}         & \usym{2713}         &                 & \usym{2713}            & \usym{2713}                                                                      & 74.7          & 30.9          & 21.7          & 23.5          & 47.6          & 64.9          \\ \midrule
\multicolumn{1}{l|}{Ours}                    & D          & \usym{2713}         & \usym{2713}         & \usym{2713}               &              & \usym{2713}                                                                      & \textbf{86.1} & \textbf{77.1} & \textbf{55.0} & \textbf{59.8} & \textbf{73.7} & \textbf{83.7} \\ \bottomrule [0.8pt]

\end{tabular}
\end{table*}

\subsection{Experimental Setup}

\textbf{Datasets.}
The REAL275 and CAMERA25 datasets were proposed by the creators of NOCS~\cite{wang2019normalized}. Both datasets consist of six different categories: \emph{bottle, bowl, camera, can, laptop} and \emph{mug}. The CAMERA25 dataset contains a total of 300K synthetic RGB-D images generated by rendering and compositing virtual instances from ShapeNetCore~\cite{chang2015shapenet}. Approximately 90\% of the images, covering $1,085$ instances, are divided into the training dataset while the remaining 10\% (25K) covering $184$ different instances are reserved for evaluation. The REAL275 dataset contains real-world RGB-D images, with $4,300$ images for training and $2,750$ images for evaluation. The training set is collected from seven scenes, and the testing set contains 6 scenes with 3 unseen instances per category, each containing $18$ real objects spanning the six categories.
The YCB-Video dataset~\cite{xiang2017posecnn} is a challenging dataset characterized by strong occlusion, clutter and several symmetric objects. It contains 21 objects and 92 video sequences. The Wild6D dataset~\cite{ze2022category} is a large-scale RGB-D dataset, including 5,166 videos covering 1722 different object instances across 5 categories. We employ the same train and test sets as~\cite{ze2022category}. 

Our self-built dataset, termed "\emph{DYNAMIC45"}, was collected using an air-ground robotic system to capture RGB-D data in dynamic scenes~(see left of Fig.~\ref{FIG_vision-our}).
These data were captured by a RGB-D camera sensor (RealSense D435i) and contains 16 videos with 8 scenes and 10 categories, totaling 45K images. We split these videos into 12 for training and 4 for testing. To automatically annotate these frames, we first calibrate the camera extrinsic matrix for each video and then use an offline 3D labeling tool to obtain the necessary annotations. This dataset will be made public.

\textbf{Metrics.}
Following~\cite{irshad2022centersnap}, we evaluate the performance of 6-DoF pose/size estimation and 3D shape recovery using four types of metrics: (1) \emph{3D Intersection over Union} (3D-IoU), measures the average accuracy for various IoU-overlap thresholds (we take 50\% and 75\% as the thresholds). (2) \emph{${a^ \circ }b~cm$}, measures the fraction of the rotation error less than $a$ and the translation error less than $b$ for each instance~(we take ${5^ \circ }2~cm$, ${5^ \circ }5~cm$, ${10^ \circ }2~cm$ and ${10^ \circ }5~cm$ as the thresholds). (3) On YCB-Video dataset, we employ extra commonly used metrics, including ADD (S) and the AUC (are under curve) of ADD-S and ADD(-S), similar to~\cite{wang2021occlusion}. (4) For 3D shape reconstruction, we adopt the \emph{Clamfer Distance} (CD), that calculates the average closest point distance, following \cite{tian2020shape}.

\textbf{Implementation Details.}
Our network is implemented using PyTorch and trained in REAL275 and CAMERA25. We train our model using the Adam Optimizer with initial learning rate of ${\rm{1}}{{\rm{e}}^{ - 4}}$ (decayed by 0.7 every 40 epochs) and a batch size of 32 on an NVIDIA RTX A6000 GPU. To ensure fairness, we also utilize the available segmentation results generated by an off-the-shelf network (Mask-RCNN) to crop all objects from the RGB-D frame by using camera intrinsic parameters and recover the partial observable point cloud ${P_0}$. We pre-sample the points in ${D_{data}}({P_0})$ such that ${N_0} = N$. The number of points in each shape prior is set as $2,048$. The number of timesteps ${T}$ in the diffusion process is $100$, and we set the variance schedules from ${\beta _{\rm{1}}} = {\rm{1}}{0^{ - 4}}$ to ${\beta _{\rm{T}}} = 0.5 \times {\rm{1}}{0^{ - {\rm{1}}}}$, with ${\beta _{\rm{t}}}$ being linearly interpolated.

\subsection{Comparison and Evaluation on Public Datasets}

\subsubsection{Fully-Supervised Category-Level Pose Estimation}
\begin{figure*}[ht]
    \begin{minipage}[t]{0.5\linewidth}
        \centering
        \includegraphics[width=0.9\textwidth]{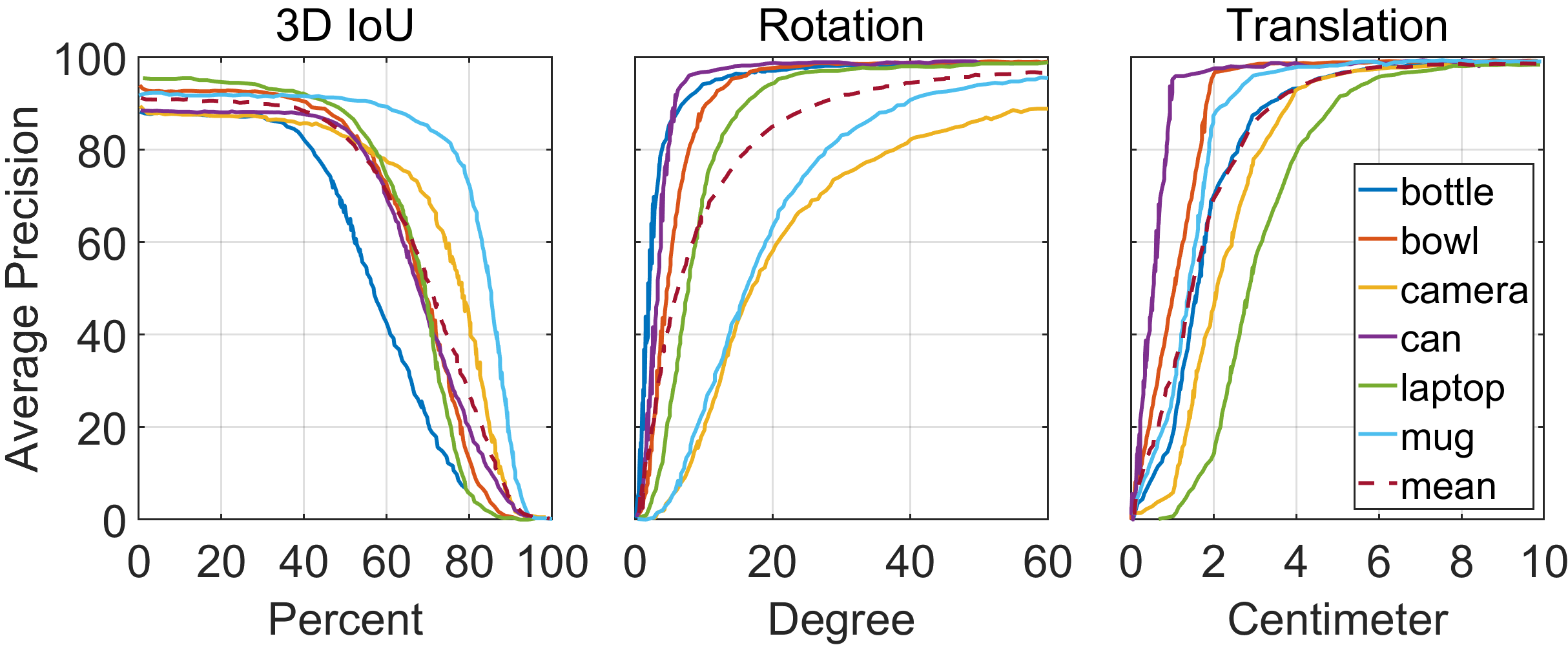}
        \centerline{(a) Baseline (NOCS) results on CAMERA25 dataset.}
    \end{minipage}%
    \vspace{0.2cm}
    \begin{minipage}[t]{0.5\linewidth}
        \centering
        \includegraphics[width=0.9\textwidth]{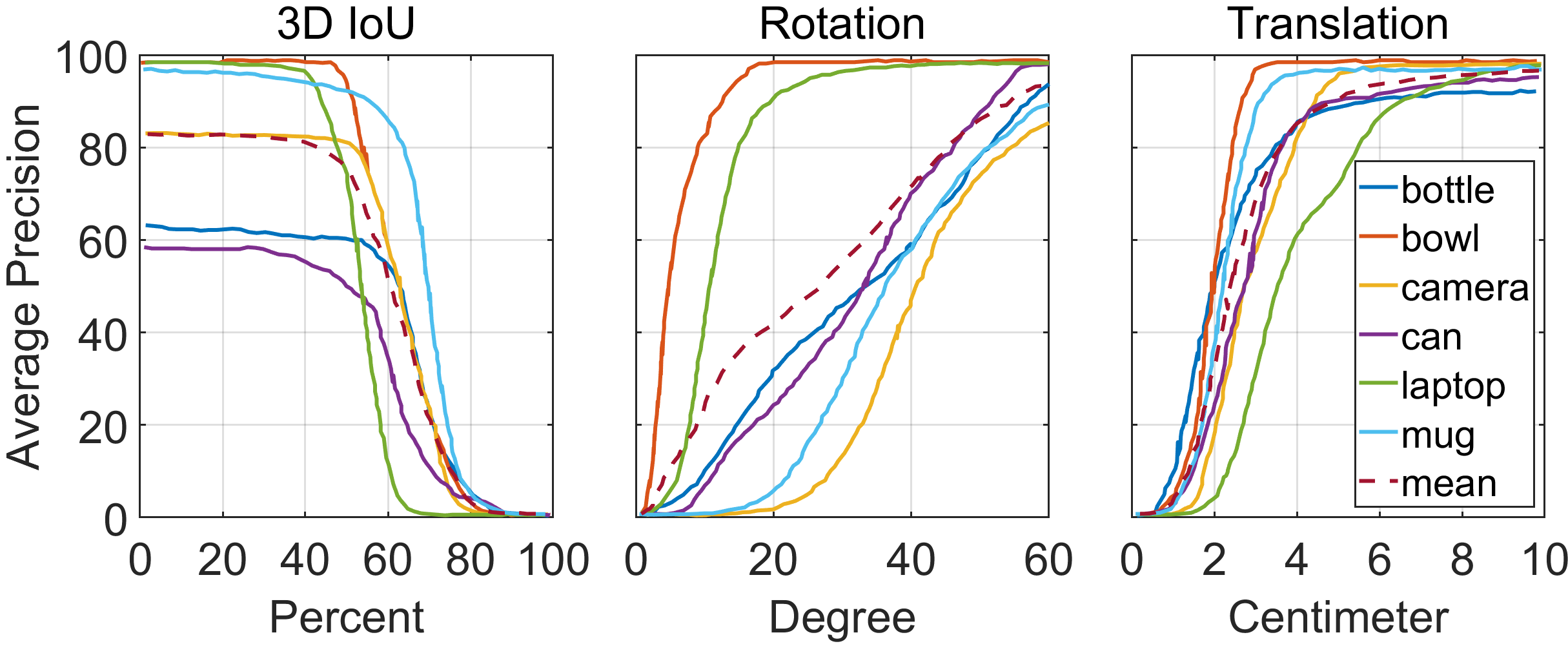}
        \centerline{(b) Baseline (NOCS) results on REAL275 dataset.}
    \end{minipage}

    \begin{minipage}[t]{0.5\linewidth}
        \centering{}
        \includegraphics[width=0.9\textwidth]{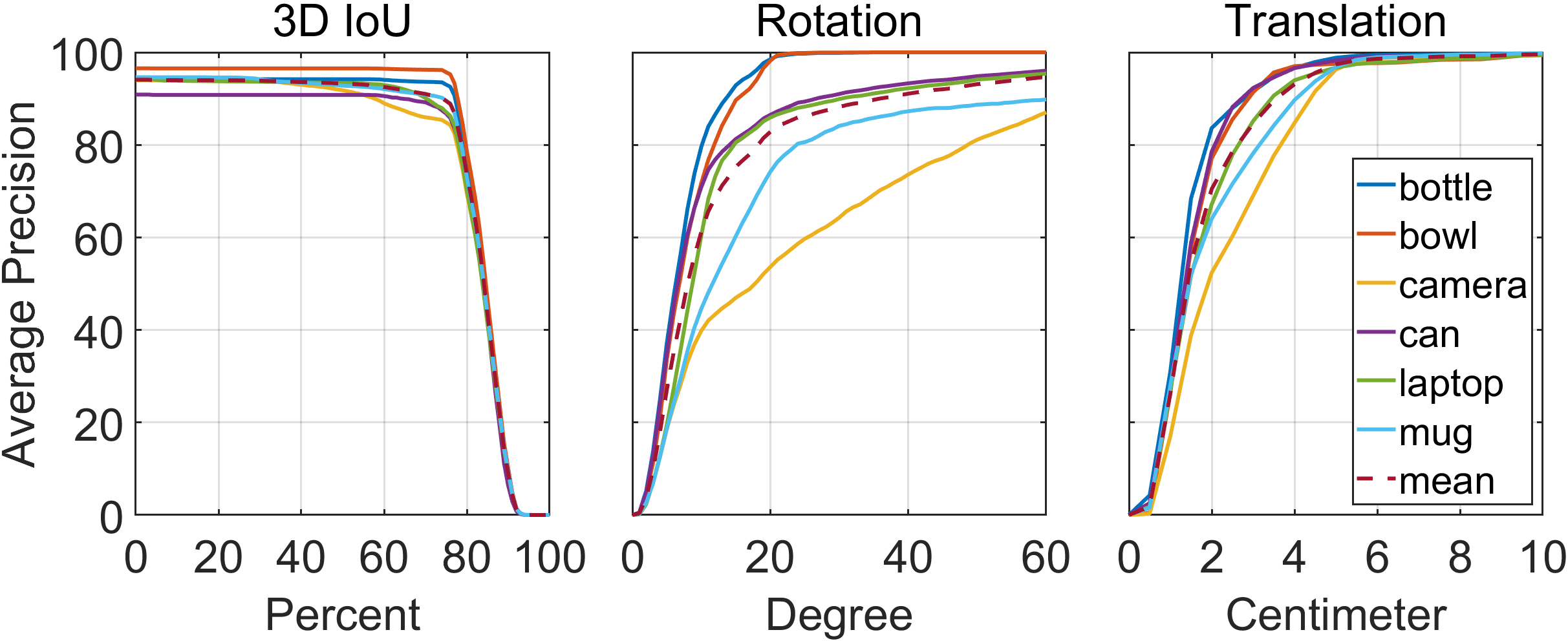}
        \centerline{(c) Our results on CAMERA25 dataset.}
    \end{minipage}%
    \vspace{0.2cm}
    \begin{minipage}[t]{0.5\linewidth}
        \centering
        \includegraphics[width=0.9\textwidth]{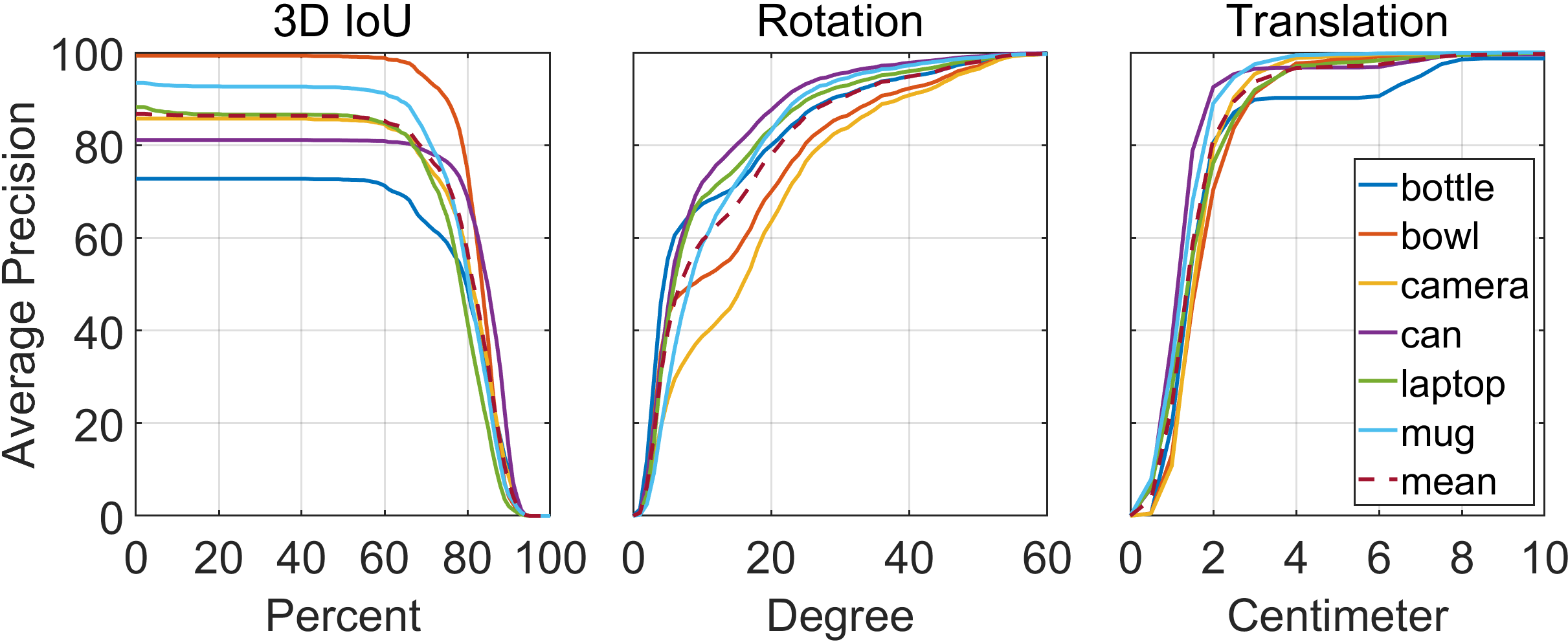}
        \centerline{(d) Our results on REAL275 dataset.}
    \end{minipage}
    \vspace{-0.3cm}
    \caption{\textbf{Comparison of mAP on both CAMERA25 and REAL275 datasets.} Mean average percision (mAP) of our method and baseline for various 3D IoU, rotation and translation thresholds on CAMERA25 and REAL275 datasets. The upper row (\emph{i.e.,} (a) and (b)) is the results of NOCS and all results are from original paper~\cite{wang2019normalized}, and the bottom row (\emph{i.e.,} (c) and (d)) is ours.}
    \label{fig-map}
\end{figure*}

We first compare the performance of our approach with state-of-the-art category-level pose estimation methods trained under the full supervision of pose annotations. As shown in Table~\ref{table_full}, we compare our results with $14$ available full-supervised methods on both CAMERA25 and REAL275 datasets. Our final refined model achieves state-of-the-art results for all evaluation metrics on both two dataset. Notably, on the CAMERA25 dataset, we outperforms Query6DoF~\cite{wang2023query6dof}, the current SoTA method, by 4.0\% and 4.6\% in terms of $IoU75$ and ${10^ \circ }2{\rm{cm}}$. Similarly, on another real-world REAL275 dataset, we also outperform other available baselines. Specifically, our method exhibits superior performance, achieving an mAP of 86.1\% for $IoU50$, 55.0\% for ${5^ \circ }2{\rm{cm}}$ and 59.8\% for ${5^ \circ }5{\rm{cm}}$. These results indicate that the robustness performance of our diffusion-based self-supervised learning.
Furthermore, we report comparative results of the category-specific mAP (mean average percision) between our method and NOCS~\cite{wang2019normalized} across both public datasets. Detailed results can be found in Fig.~\ref{fig-map}, where our method outperforms the NOCS by a large margin across most categories in both datasets. These experiments show that our method's superior performance for category-level pose estimation.

\subsubsection{Self-Supervised Category-Level Pose Estimation}
\begin{figure*}[ht]
    \centering
    \includegraphics[width=\textwidth]{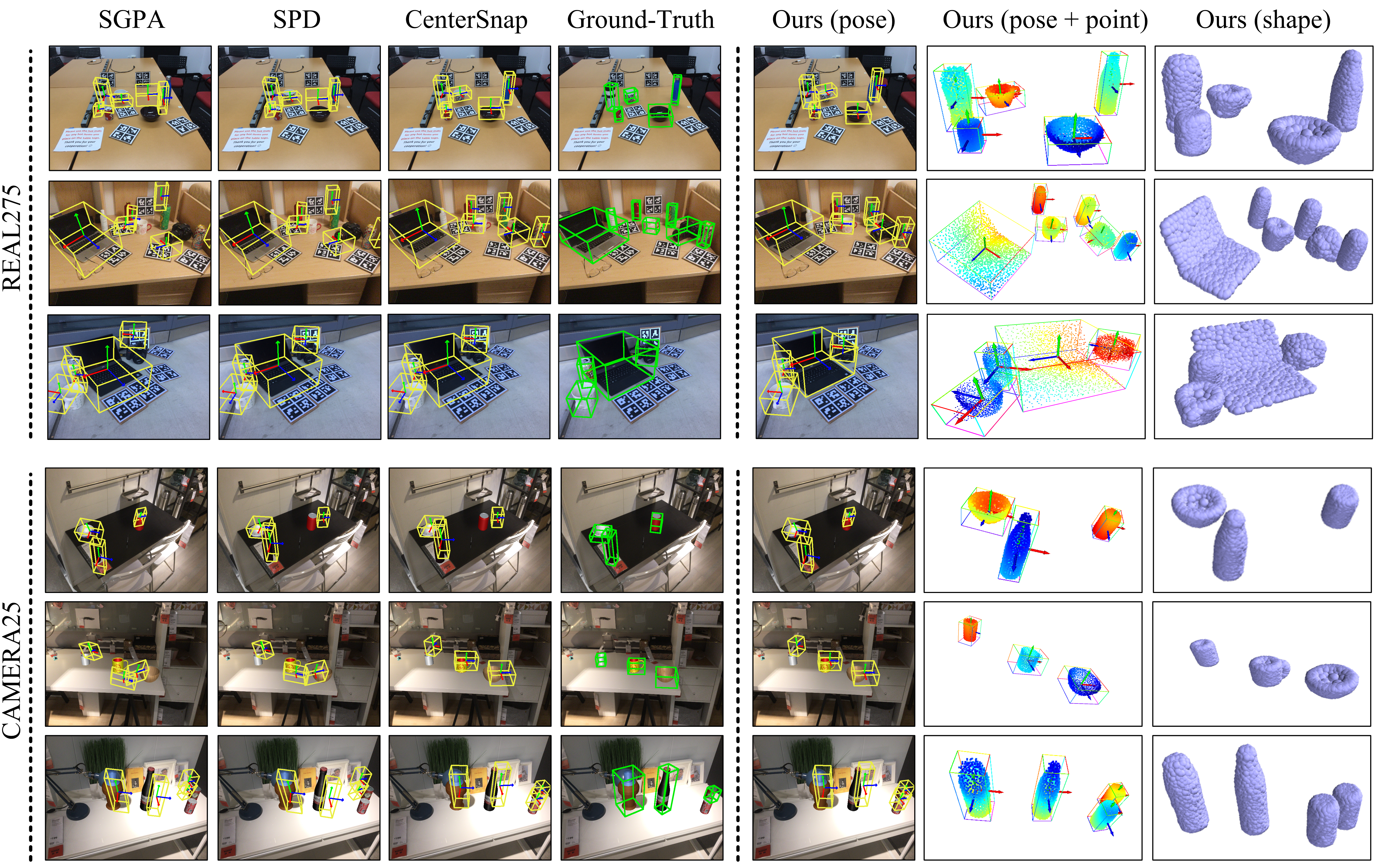}   
    \caption{\textbf{Visualization comparison results on both REAL275 and CAMERA25 datasets.} We compare our method with other representative SoTA approaches in terms of 6-DoF pose and size estimation. Meanwhile, we also show the corresponding result of 3D shape reconstruction (right side). The yellow and green boxes represent the predictions and ground-truth labels, respectively.}\label{FIG_vision-real-syn}
    \vspace{-0.2cm}
\end{figure*}

We conducted some experiments to compare our method with existing published category-level self-supervised methods on REAL275 dataset, including those without the use of shape prior~\cite{lee2022uda,zaccaria2023self} and those utilizing shape prior~\cite{peng2022self,lin2022category,zhang2022self,yu2023robotic}. It's worth noting that our method is currently the only one for multi-object pose estimation among these self-supervised approaches. As depicted in TABLE~\ref{table_self}, we outperform others by a significant margin across all evaluation metrics. In detail, we achieve 55.0\% and 59.8\% on ${5^ \circ }2{\rm{cm}}$ and ${5^ \circ }5{\rm{cm}}$, respevtively, surpassing Self-DPDN~\cite{lin2022category}, the current most powerful method, by 15.6\% and 14.8\%. Regarding prior-free methods, compared with UDA-COPE~\cite{lee2022uda}, we demonstrate substantial improvements across all metrics, \emph{e.g.,} 77.1\% vs. 62.5\% on $IoU75$, 55.0\% vs. 30.4\% on ${5^ \circ }2{\rm{cm}}$ and 59.8\% vs. 34.8\% on ${5^ \circ }5{\rm{cm}}$. Qualitative visualization comparisons between several state-of-the-art (SoTA) approaches and our method across different scenes in REAL275 and CAMERA25 dataset are shown in Fig.~\ref{FIG_vision-real-syn}.

\begin{table}[ht]
\scriptsize
\footnotesize
\setlength{\tabcolsep}{0.5pt}
\centering
\renewcommand\arraystretch{1.0}
\caption{\textbf{Quantitative comparison of category-level pose estimation on the pubilc Wild6D dataset.} Note that the results of SSC-6D~\cite{peng2022self} and Self-DPDN~\cite{lin2022category} are from our test using their public codes, and other results are from Self-Pose~\cite{zhang2022self}. Column of Data indicates the training data: "C=CAMERA25", "R=REAL275" and "W=Wild6D". "Full." and "Self." represent the learning scheme.}
\label{table_wild6D}
\begin{tabular}{l|c|c|ccccc}
\toprule [0.8pt] 
\multicolumn{1}{c|}{\multirow{2}{*}{Method}}                & \multirow{2}{*}{Scheme} & \multirow{2}{*}{Data} & \multicolumn{5}{c}{Evaluation Metrics} \\
\multicolumn{1}{c|}{}                                       &                         &                       & $IoU50$  & ${5^ \circ }2{\rm{cm}}$ & ${5^ \circ }5{\rm{cm}}$ & ${10^ \circ }2{\rm{cm}}$ & ${10^ \circ }5{\rm{cm}}$ \\ \midrule
CASS~\cite{chen2020learning}                                                                    & \multirow{5}{*}{Full.}  & C+R                   & 1.0    & 0     & 0     & 0     & 0     \\
SPD~\cite{tian2020shape}                                                                        &                         & C+R                   & 32.5   & 2.6   & 3.5   & 9.7   & 13.9  \\
DualNet~\cite{lin2021dualposenet}                                                               &                         & C+R                   & 70.0   & 17.8  & 22.8  & 26.3  & 36.5  \\
GPV-Pose~\cite{di2022gpv}                                                                       &                         & C+R                   & 67.8   & 14.1  & 21.5  & 23.8  & 41.1  \\
RePoNet~\cite{ze2022category}                                                                   &                         & C+W                   & \underline{70.3}   & 29.5  & 34.4  & 35.0  & 42.5  \\ \midrule
SSC-6D~\cite{peng2022self}                                                                      & \multirow{3}{*}{Self.}  & C+R                   & 68.2   & 25.2  & 30.9  & 33.9  & 39.6  \\
Self-DPDN~\cite{lin2022category}                                                                &                         & C+R                   & 69.5   & 31.6  & \underline{36.3}  & \underline{40.0}  & 43.2  \\
Self-Pose~\cite{zhang2022self}                                                                  &                         & W                     & 68.2   & \underline{32.7}  & 35.3  & 38.3  & \underline{45.3}  \\ \midrule
Ours                                                                                            & \multirow{2}{*}{Self.}  & C+R                   & \textbf{71.1}   & \textbf{33.3}  & \textbf{38.5}  & \textbf{43.6}  & \textbf{48.4}  \\
Ours (+ Wild6D)                                    &                         & C+R+W                 & \textbf{73.0}   & \textbf{35.2}  & \textbf{39.7}  & \textbf{46.4}  & \textbf{50.9}  \\ \bottomrule [0.8pt]
\end{tabular}\vspace{-0.2cm}
\end{table}

\begin{table}[ht]
\scriptsize
\footnotesize
\setlength{\tabcolsep}{3.5pt}
\centering
\renewcommand\arraystretch{1.0}
\caption{\textbf{Quantitative comparison of instance-level pose estimation on the pubilc YCB-Video dataset.} The comparison results of other baselines are summarized from their original papers, and empty entries indicate that they were not reported. Column of Supervision indicates the training manner: "Syn.=synthetic data", "Real=real-world data" and "GT=ground-truth". "Semi." and "Self." represent the learning scheme.}
\label{table_ycb}
\begin{tabular}{l|c|ccc}
\toprule [0.8pt]
\multicolumn{1}{c|}{\multirow{2}{*}{Method}} & \multirow{2}{*}{Supervision}    & \multicolumn{3}{c}{Evaluation Metrics}                                                                                            \\
\multicolumn{1}{c|}{}                        &                                 & ADD (S)        & \begin{tabular}[c]{@{}c@{}}AUC of\\ ADD-S\end{tabular} & \begin{tabular}[c]{@{}c@{}}AUC of\\ ADD (-S)\end{tabular} \\ \midrule
PoseCNN~\cite{xiang2017posecnn}                                      & \multirow{5}{*}{Syn. + Real GT} & 21.3          & 75.9                                                   & 61.3                                                     \\
PVNet~\cite{peng2019pvnet}                                        &                                 & -             & -                                                      & 73.4                                                     \\
DeepIM~\cite{li2018deepim}                                       &                                 & -             & 88.1                                                   & 81.9                                                     \\
CosyPose~\cite{labbe2020cosypose}                                     &                                 & -             & 89.8                                                   & \underline{84.5}                                               \\
SO-Pose~\cite{di2021so}                                      &                                 & 56.8          & 90.9                                             & 83.9                                                     \\ \midrule
Zhou et al.~\cite{zhou2021semi}                                  & Real+ Semi.                     & -             & 90.4                                                   & -                                                        \\
Self6D++~\cite{wang2021occlusion}                                     & Syn. + Self.                    & -             & \underline{91.1}                                                   & 80.0                                                     \\
Hai et al.~\cite{hai2023pseudo}                                   & Real+ Self.                     & \underline{67.4}    & -                                                      & -                                                        \\ \midrule
Ours                                         & Syn.+ Real+ Self.               & \textbf{70.7} & \textbf{92.0}                                          & \textbf{85.3}                                            \\ \bottomrule [0.8pt]
\end{tabular}
\end{table}

\begin{figure*}[!t]
\centering 
\subfigure[YCB-Video dataset.]{
\includegraphics[width=6.5cm,height = 5cm]{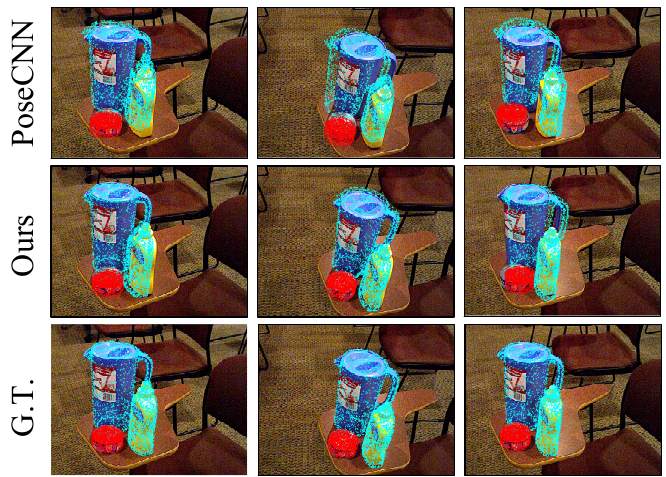}}
\subfigure[Wild6D dataset.]{
\includegraphics[width=11.3cm,height = 5cm]{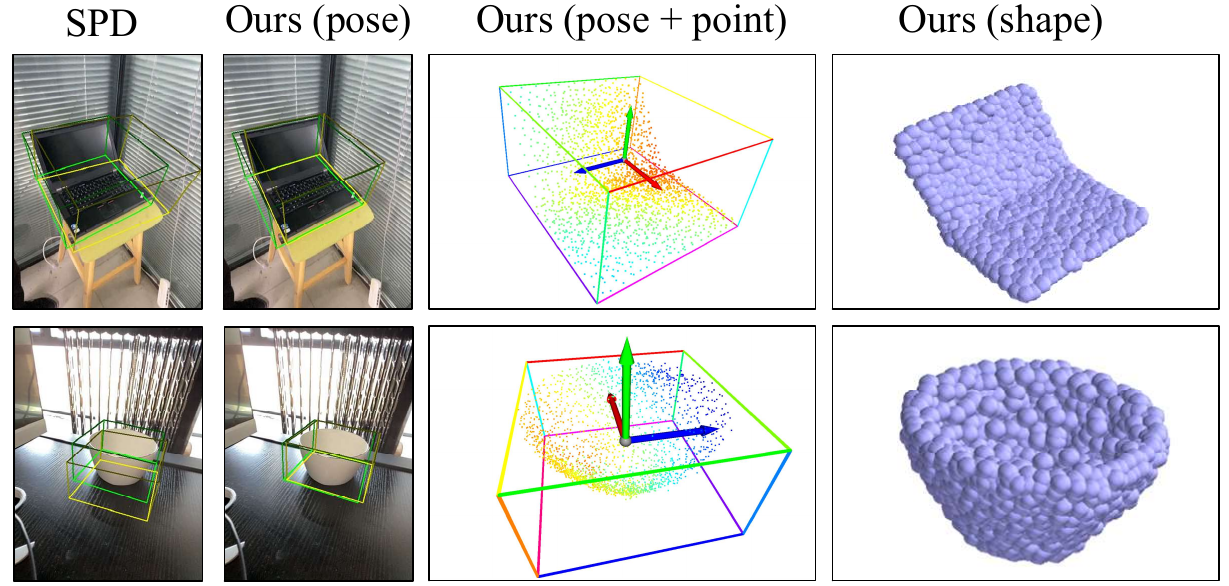}}
\caption{\textbf{Visualization comparison results on both YCB-Video and Wild6D datasets.} (a) Qualitative results of our method and PoseCNN~\cite{xiang2017posecnn} on YCB-Video dataset. To keep in line with PoseCNN, each object shape model are transformed with the predicted pose and then projected into the 2D images. (b) Qualitative results of our method and SPD~\cite{tian2020shape} on Wild6D dataset. Yellow and green represent predicted and ground-truth results.}\label{FIG_vision_ycb-wild}
\vspace{-0.2cm}
\end{figure*}

Moreover, to assess our method's generalization performance, we present the results on the public Wild6D dataset. We compare with both fully-supervised and self-supervised category-level methods by using our trained model on REAL275 and CAMERA25 and our model with fine-tuning on Wild6D dataset, respectively. The results are reported in Table~\ref{table_wild6D}. It is evident that our unfine-tuned model outperforms other available baselines, and the coresponding fine-tuned model exhibits an obvious improvement across various metrics. Therefore, we claim that our proposed algorithm is more effective due to its utilization of a diffusion-based self-learning scheme with the SE(3)-equivariant and scale-invariant representation. Fig.~\ref{FIG_vision_ycb-wild} (b) presents some qualitative results on Wild6D.

\begin{table*}[]
\scriptsize
\footnotesize
\setlength{\tabcolsep}{3pt}
\centering
\renewcommand\arraystretch{1.0}
\caption{\textbf{Quantitative comparison of 3D shape reconstruction on pubilc CAMERA25 and REAL275 dataset: Evaluated with \emph{CD} metric (${\rm{1}}{0^{ - 2}}$).} Note that the best and the second best results are highlighted in \textbf{bold} and \underline{underlined}, and the lower the better. The comparison results of current state-of-the-art baselines are summarized from their original papers. Empty entries were not reported in the original paper.}
\label{table_shape}

\begin{tabular*}{\textwidth}{@{\extracolsep{\fill}}cl|ccccccc|ccccccc}
\toprule [0.8pt] 

\multicolumn{2}{c|}{\multirow{2}{*}{Method}}                                                               & \multicolumn{7}{c|}{CAMERA25 Dataset}                                                                                                                     & \multicolumn{7}{c}{REAL275 Dataset}                                                                                                                       \\  
\multicolumn{2}{c|}{}                                                                                      & Bottle       & Bowl      & Camera       & Can          & Laptop       & Mug          & Mean         & Bottle       & Bowl      & Camera       & Can          & Laptop       & Mug          & Mean         \\ \midrule
1                 & CASS~\cite{chen2020learning}~{\color{gray}[CVPR2020]}                                                                                   & -                     & -                  & -                     & -                     & -                     & -                     & -                     & 0.17                  & 0.09                  & 0.53                  & 0.18                  & 0.19                  & 0.24                  & 0.23                  \\
2                 & Reconstruction~\cite{tian2020shape}~{\color{gray}[ECCV2020]}                                                                            & 0.18                  & 0.16                  & 0.40                  & 0.09                  & 0.20                  & 0.14                  & 0.20                  & 0.34                  & 0.12                  & 0.89                  & 0.15                  & 0.29                  & \underline{0.10}                  & 0.32                  \\
3                 & SPD~\cite{tian2020shape}~{\color{gray}[ECCV2020]}                                                                            & 0.17                  & 0.15                  & 0.42                  & 0.09                  & 0.20                  & 0.14                  & 0.20                  & 0.34                  & 0.12                  & 0.89                  & 0.15                  & 0.29                  & \underline{0.10}                  & 0.32                  \\
4                 & C3R-Net~\cite{wang2021category}~{\color{gray}[IROS2021]}                                                                              & 0.13                     & \textbf{0.10}                  & 0.26                     & \underline{0.08}                     & \underline{0.09}                     & \underline{0.11}                     & \underline{0.13}                     & 0.30                     & 0.10                  & 0.76                     & 0.13                   & 0.13                   & 0.12                   & 0.26                   \\
5                 & SGPA~\cite{chen2021sgpa}~{\color{gray}[ICCV2021]}                                                                             & 0.13                  & 0.13                  & 0.33                  & \underline{0.08}                  & 0.12                  & \underline{0.11}                  & 0.15                  & 0.29                  & 0.09                  & 0.55                  & 0.17                  & 0.16                  & 0.11                  & 0.23                  \\
6                 & CenterSnap~\cite{irshad2022centersnap}~{\color{gray}[ICRA2022]}                                                                             & \underline{0.11}                  & \textbf{0.10}                  & 0.29                  & 0.13                  & \textbf{0.07}                  & 0.12                  & 0.14                  & 0.13                  & 0.10                  & 0.43                  & 0.09                  & \underline{0.07}                  & \textbf{0.06}                  & 0.15                  \\ 
7                 & SSC-6D~\cite{peng2022self}~{\color{gray}[AAAI2022]}                                                                             & -                  & -                  & -                  & -                  & -                  & -                  & -                  & \textbf{0.06}                  & \textbf{0.04}                  & 0.38                  & 0.10                  & \textbf{0.03}                  & 0.45                  & 0.18                  \\
8                 & 6D-ViT~\cite{zou20226d}~{\color{gray}[TIP2022]}                                                                                   & 0.14                  & \underline{0.12}                  & 0.29                  & 0.09                  & 0.12                  & 0.12                  & 0.15                  & 0.24                  & 0.11                  & 0.61                  & 0.16                  & 0.14                   & 0.11                  & 0.21                  \\
9                 & ShAPO~\cite{Irshad2022ShAPOIR}~{\color{gray}[ECCV2022]}                                                                           & 0.14                  & 0.13                  & \underline{0.20}                  & 0.14                  & \textbf{0.07}                  & \underline{0.11}                  & 0.16                  & \underline{0.10}                  & 0.08                  & 0.40                  & \underline{0.07}                   & 0.08                   & \textbf{0.06}                  & \underline{0.13}                  \\ 
10                 & GCASP~\cite{li2023generative}~{\color{gray}[PMLR2023]}                                                                             & -                  & -                  & -                  & -                  & -                  & -                  & -                  & 0.21                  & 0.16                  & \underline{0.11}                  & 0.16                  & 0.21                  & 0.29                  & 0.19                  \\ \midrule
                  \multicolumn{2}{l|}{Ours}                                                                          & \textbf{0.10}                  & \textbf{0.10}                  & \textbf{0.17}                  & \textbf{0.07}                  & \textbf{0.07}                  & \textbf{0.08}                  & \textbf{0.10}                  & \textbf{0.06}                  & \underline{0.07}                  & \textbf{0.09}                  & \textbf{0.06}                  & \textbf{0.03}                  & \textbf{0.06}                  & \textbf{0.06}                  \\ \bottomrule [0.8pt]
\end{tabular*}
\end{table*}

\begin{figure*}[!t]
    \centering
    \subfigure[CAMERA25 dataset.]
    {    \centering
        \includegraphics[width=0.48\textwidth]{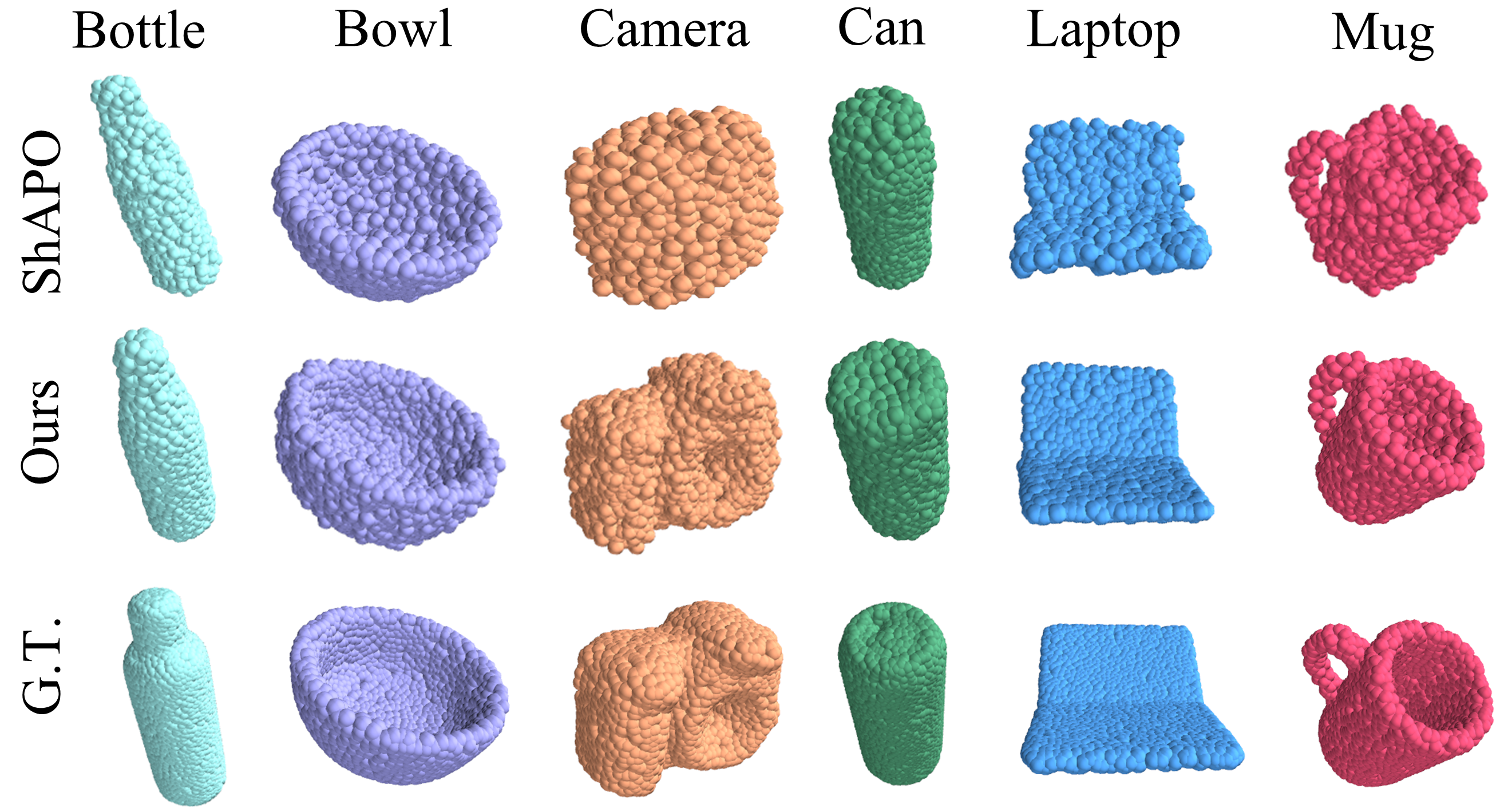}
                
    }
    \subfigure[REAL275 dataset.]
    {    \centering
        \includegraphics[width=0.48\textwidth]{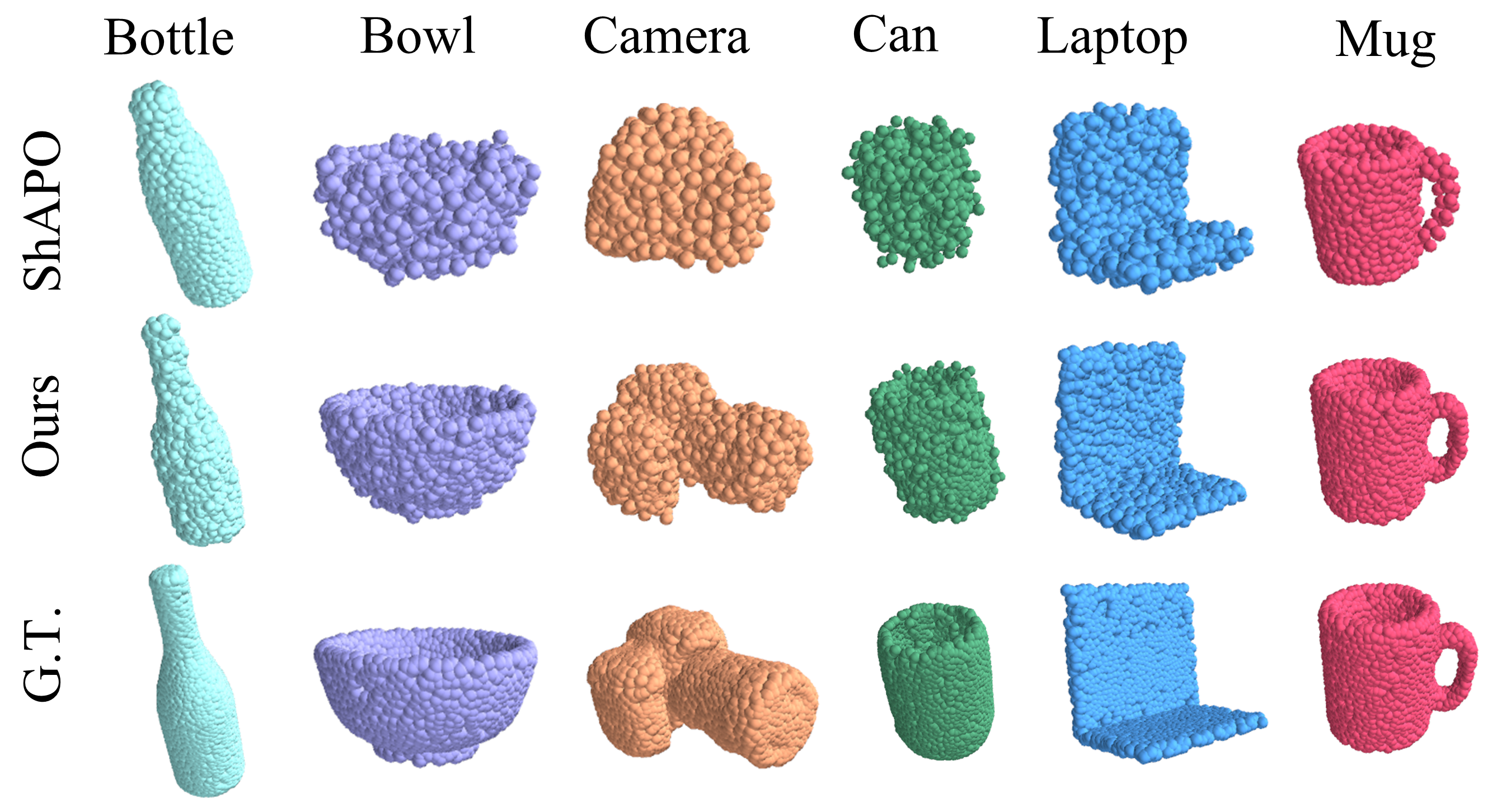}
           
    }
    \caption{\textbf{Visualization comparison results of 3D shapes reconstruction.} We compared our proposed method with the strong currently available baseline (ShAPO~\cite{Irshad2022ShAPOIR}) for all six categories on both CAMERA25 and REAL275 datasets. It is clear that our results are closer to the ground-truth. To facilitate comparison, we align all results with same category into a similar pose.}\label{fig_shape-compare}
\end{figure*}

\begin{table}[]
\scriptsize
\footnotesize
\setlength{\tabcolsep}{1pt}
\centering
\renewcommand\arraystretch{1.0}
\caption{\textbf{Quantitative comparison of 3D shape reconstruction on pubilc YCB-Video dataset: Evaluated with \emph{CD} metric (${\rm{1}}{0^{ - 2}}$).} Note that all reported results are from models using their public codes or checkpoints and fine-tuned on YCB-Video for a fair comparison.}
\label{table_shape_ycb}

\begin{tabular}{l|ccc|cc}
\toprule [0.8pt]
\multicolumn{1}{c|}{\multirow{2}{*}{Object}} & \multicolumn{3}{c|}{Fully-Supervised} & \multicolumn{2}{c}{Self-Supervised} \\  
\multicolumn{1}{c|}{}                        & SGPA~\cite{chen2021sgpa}      & SPD~\cite{tian2020shape}      & ShAPO~\cite{Irshad2022ShAPOIR}     & SSC-6D~\cite{peng2022self}            & Ours            \\ \midrule 
004 sugar box                                & 0.15      & 0.20     & \underline{0.11}           & \underline{0.11}              & \textbf{0.10}            \\
019 pitcher base                             & 0.24      & 0.31     & \underline{0.17}           & 0.18              & \textbf{0.14}            \\
021 bleach cleanser                          & 0.30      & 0.33     & 0.27           & \underline{0.20}              & \textbf{0.17}            \\ 
036 wood block                               & \underline{0.09}      & 0.11     & \underline{0.09}           & \textbf{0.07}              & \textbf{0.07}            \\
040 large marker                             & \underline{0.14}      & 0.18     & 0.22           & 0.16              & \textbf{0.10}            \\
\midrule
\multicolumn{1}{l|}{Mean}                    & 0.18      & 0.23     & 0.17           & \underline{0.14}              & \textbf{0.12}            \\ \bottomrule [0.8pt]
\end{tabular}
\vspace{-0.5cm}
\end{table}
\subsubsection{Self-Supervised Instance-Level Pose Estimation}
Table~\ref{table_ycb} presents a comparison of our method with various instance-level approaches on YCB-Video dataset, using the common metric of ADD(S) and the extra standard metric AUC of ADD-S/ADD(-S). 
We compare not only with available instance-level approaches under the fully-supervision setting (row-1 to row-5 in Table~\ref{table_ycb}), but also with state-of-the-art baselines under the self-supervised~\cite{wang2021occlusion,hai2023pseudo} or semi-supervised~\cite{zhou2021semi} manner. Although our method is not specifically designed for these instance-level self-supervised pose estimation, such as Self6D++~\cite{wang2021occlusion}, our model still achieves robust performance, compared to or even surpassing the state-of-the-arts. Notably, our approach shows improved results for the ADD(S) metric with 70.7\% compared to 67.4\% reported in~\cite{hai2023pseudo}, and achieves slightly better performance for the AUC of ADD-S and ADD(-S) with 92.0\% vs. 91.1\% and 85.3\% vs. 80.0\% from Self6D++~\cite{wang2021occlusion}. We also provide  qualitative results on YCB-Video dataset, displayed in Fig.~\ref{FIG_vision_ycb-wild} (a).

\subsubsection{3D Shape Reconstruction Evaluation}
To evaluate the quality of shape reconstruction, we first compare our method with 10 state-of-the-art baselines using the metric of \emph{CD}, that calculates the distance between our reconstructed shape model $P$ and the ground truth shape model $\hat P$ in the canonical space.
As reported in Table~\ref{table_shape}, our method obtains average \emph{CD} metrics of 0.10 on the CAMERA25 dataset and 0.06 on the REAL275 dataset, outperforming the current well-performed baseline, ShAPO~\cite{Irshad2022ShAPOIR} by 0.06 and 0.07, respectively. Notably, among these baselines, SSC-6D~\cite{peng2022self} is the only self-supervised approach, and our model outperforms it across the majority of categories. The main reason for this case is that these baselines tend to produce a mean shape for each instance within the same category, however, our method tries to seek the complete shape from partially observable point clouds, that may be closer to the ground-truth than the mean shape. This distinction is evident in the visualization comparison results displayed in Fig.~\ref{fig_shape-compare}. 

Furthermore, to verify our proposed method's shape reconstruction performance for different shape categories, we conduct an additional experiment on the YCB-Video dataset, considering five new unseen object categories such as box, etc. As depicted in Table~\ref{table_shape_ycb}, we compared our method with several fully-supervised approaches (SGPA~\cite{chen2021sgpa}, SPD~\cite{tian2020shape} and ShAPO~\cite{Irshad2022ShAPOIR}), as well as the currently available self-supervised baseline~SSC-6D~\cite{peng2022self}, respectively. On this benchmark dataset, our method also achieves the state-of-the-art results.
From these comparative results and the visualization of category-specific shape reconstruction, it is evident that our self-supervised network effectively enhances the quality of 3D shape reconstruction. This further indicates that our proposed diffusion-driven pretrain-to-refine self-supervised learning strategy not only can improve the performance of pose and 3D size estimation, but also facilitates corresponding 3D shape reconstruction.

\subsection{Ablation Studies and Analysis}
\subsubsection{Evaluation of Different Components of ORT}
To validate the effectiveness of the specific component design in our ORT module, we conduct the detailed ablation experiments on REAL275 dataset, as depicted in Table.~\ref{table_abl-com}. The last row in the table represents the complete ORT module, while the first three rows display the results of ablating the point convolution layer with radial-kernels (RK-Point layer), the 3D scale-invariant graph convolution layer (3DSI-Graph layer) and the SE(3) group convolution layer (SE(3)-Group layer), respectively. 
Additionally, we examined the effect of our seed points aggregation and the results using the base Transformer model without adding any proposed components (w/o SE(3) block).  

\emph{1) Effect of RK-Point layer:} Initially, we investigate the impact of the proposed RK-Point layer on the model performance by removing this layer while remaining other layers. As shown in the row-1 in Table.~\ref{table_abl-com}, no RK-Point Layer resultes in a significant decline in most pose estimation metrics upon ablation, \emph{i.e.,} $IoU50$ : 25.6\%$\downarrow$, ${5^ \circ }2{\rm{cm}}$ : 15.1\%$\downarrow$ and ${10^ \circ }5{\rm{cm}}$ : 18.4\%$\downarrow$, while the shape reconstruction metric remains essentially unchanged. This proves the effectiveness of our radial kernel mapping in selecting SE(3) equivariant pose-aware features.

\emph{2) Effect of 3DSI-Graph layer:} Next, we verify the effectiveness of our 3DSI-Graph Layer using the same operation as above. The results from row-2 in Table.~\ref{table_abl-com} demonstrate a decrease of 0.18 in the CD metric upon ablating the 3DSI-Graph Layer, with minimal decay observed in pose metrics. 
This confirms the effectiveness of our proposed scale-invariant graph convolution layer in enhancing the network's ability to capture scale/shape invariance information for shape reconstruction.

\emph{3) Effect of SE(3)-Group layer:} The third experiment aims to verify the effectiveness of the SE(3) group convolution layer. By comparing the results in row-3 and row-6 of Table.~\ref{table_abl-com}, we observed that incorporating this layer into ORT module further improves both pose estimation and shape reconstruction, \emph{e.g.,} ${10^ \circ }5{\rm{cm}}$ : 13.7\%$\uparrow$ and CD : 0.04$\downarrow$. This highlights the effectiveness of the SE(3)-Group Layer. 

\emph{4) Effect of SE(3) block:} The result in row-4 reveals the performance of our simplest model without adding any proposed components, \emph{w.r.t.,} the ORT module is a modification of multi-head attention. It's clear to see that incorporating our complete SE(3) block into the ORT module drastically improves the model performance across almost all metrics, compared to row-6. These results further show the important role of our proposed three layers in the effectiveness of ORT module.

\emph{5) Impact of seed points generation:} We also evaluate the impact caused by the seed points generation strategy used in ORT. It can be seen that the performance of our model slightly decline when the seed points are absent. The potential reason is that the neighbor domain space obtained from seed points can effectively establish a link between prior and observation, resulting in more relevant features.
\begin{table}[t]
\scriptsize
\footnotesize
\setlength{\tabcolsep}{1pt}
\centering
\renewcommand\arraystretch{1.0}
\caption{\textbf{Ablation study of our ORT module to investigate each components contribution on REAL275 dataset.} Note that 3D shape reconstruction is evaluated with \emph{CD} metric (${\rm{1}}{0^{ - 2}}$).}
\label{table_abl-com}

\begin{tabular}{l|ccccc|c}
\toprule [0.8pt]
\multicolumn{1}{c|}{\multirow{2}{*}{Method}} & \multicolumn{5}{c|}{Pose and Size} & Shape \\
\multicolumn{1}{c|}{}                        & $IoU50$  & ${5^ \circ }2{\rm{cm}}$  & ${5^ \circ }5{\rm{cm}}$  & ${10^ \circ }2{\rm{cm}}$ & ${10^ \circ }5{\rm{cm}}$ & CD       \\ \midrule
Ours (w/o RK-Point)                    & 60.5   & 39.9   & 49.2   & 62.7  & 65.3  & 0.07     \\
Ours (w/o 3DSI-Graph)                  & 84.6   & 52.9   & 55.3   & 71.2  & 81.6  & 0.24     \\
Ours (w/o SE(3)-Group)                 & 72.4   & 46.5   & 50.1   & 62.8  & 70.0  & 0.10     \\ 
\midrule
Ours (w/o SE(3) block)                 & 59.2   & 38.5   & 46.1   & 50.8  & 62.4  & 0.30     \\
Ours (w/o seed points)                 & 84.6   & 52.9   & 57.5   & 73.0  & 82.2  & 0.08     \\
\midrule
Ours                                         & \textbf{86.1}   & \textbf{55.0}   & \textbf{59.8}   & \textbf{73.7}  & \textbf{83.7}  & \textbf{0.06}     \\ \bottomrule [0.8pt]
\end{tabular}
\end{table}

\begin{table}[ht]
\scriptsize
\footnotesize
\setlength{\tabcolsep}{2pt}
\centering
\renewcommand\arraystretch{1.0}
\caption{\textbf{Comparison with different point sampling manners on REAL275 dataset.} Note that 'RS', 'US' and 'FPS' mean traditional random sampling, uniform sampling and farthest point sampling, respectively. 'SFD' and 'SFS' stand for the spatial feature distance and the semantic feature similarity in our SGS.}
\label{table_abl-sampling}
\begin{tabular}{l|ccccc|c}
\toprule [0.8pt]
\multicolumn{1}{c|}{\multirow{2}{*}{Sampling}} & \multicolumn{5}{c|}{Pose and Size} & Shape \\
\multicolumn{1}{c|}{}                          & $IoU50$  & ${5^ \circ }2{\rm{cm}}$  & ${5^ \circ }5{\rm{cm}}$  & ${10^ \circ }2{\rm{cm}}$ & ${10^ \circ }5{\rm{cm}}$ & CD       \\ \midrule
RS                                             & 69.7   & 30.3   & 42.1   & 53.6  & 62.2  & 0.28     \\
US                                             & 70.8   & 38.5   & 45.4   & 55.0  & 68.7  & 0.24     \\
FPS                                            & 72.2   & 40.6   & 53.7   & 62.5  & 68.9  & 0.18     \\ \midrule
SGS w/o SFD                                    & 85.0   & 48.9   & 57.5   & 70.4  & 81.6  & 0.09     \\
SGS w/o SFS                                    & 83.9   & 50.0   & 56.4   & 68.2  & 79.2  & 0.11     \\ \midrule
SGS (Ours)                                     &\textbf{86.1}   & \textbf{55.0}   & \textbf{59.8}   & \textbf{73.7}  & \textbf{83.7}  & \textbf{0.06}     \\  \bottomrule [0.8pt]
\end{tabular}
\end{table}

\begin{table}[ht]
\scriptsize
\footnotesize
\setlength{\tabcolsep}{1pt}
\centering
\renewcommand\arraystretch{1.0}
\caption{\textbf{Ablation study of proposed self-supervised training manner on REAL275 dataset.} Note that "Full" indicates our complete scheme. }
\label{table_abl-train}
\begin{tabular}{c|c|c|ccc|c}
\toprule [0.8pt]
\multirow{2}{*}{Manner}                                                        & \multirow{2}{*}{\#} & \multirow{2}{*}{Settings}  & \multicolumn{3}{c|}{Pose and Size} & Shape \\
                                                                           &                        &                            & $IoU50$      & ${5^ \circ }2{\rm{cm}}$     & ${10^ \circ }5{\rm{cm}}$     & CD    \\ \midrule
\multirow{3}{*}{\begin{tabular}[c]{@{}c@{}}Training\\ Scheme\end{tabular}} & V1                     & w/o zero conv              & 75.3       & 50.2      & 76.6      & 0.08  \\
                                                                           & V2                     & w/o trainable copy         & 66.0       & 42.3      & 58.4      & 0.10  \\
                                                                           & V3                     & directly train base model  & 65.6       & 30.1      & 52.0      & 0.13  \\ \midrule
\multirow{2}{*}{\begin{tabular}[c]{@{}c@{}}Extra\\ Condition\end{tabular}} & E1                     & no task-specific condition & 72.3       & 45.2      & 66.0      & 0.14  \\
                                                                           & E2                     & only observable latent     & 80.0       & 49.8      & 78.4      & 0.09  \\ \midrule
                                                                           & Full                 & Ours                       & \textbf{86.1}       & \textbf{55.0}      & \textbf{83.7}      & \textbf{0.06}  \\ \bottomrule [0.8pt]
\end{tabular}
\end{table}
\subsubsection{Comparison of Different Point Sampling}
In the Prior-Aware Pyramid 3D Point Transformer, we stack four shape-guided sampling modules to obtain multi-scale features and enhance the related feature representation. Here, we study the impact of different sampling manners on model performance. Specifically, we explore five schemes on the REAL275 dataset, including three traditional sampling methods and two variants of our SGS, as displayed in Table.~\ref{table_abl-sampling}. In our SGS variants (\emph{i.e.,} row-4 and row-5), all sampling points come from a single pathway. The results reveal that our SGS achieves the best performance. Relying solely on spatial feature distance or semantic feature similarity may lead to difficulties in preserving relevant points and could even result in sampling points clustering around the same instance. In this regard, we choose our proposed complete SGS approach.

\subsubsection{Effectiveness of Self-Supervised Training Manner}
To investigate the effectiveness of the proposed pretrain-to-refine self-supervised training strategy, we conduct five ablation experiments, considering two situations: training schemes and extra task-specific condition settings. 
\begin{itemize}
\item \emph{V1 "w/o zero conv"}: The zero convolution is replaced by standard convolution with Guassian weights.
\item \emph{V2 "w/o trainable copy"}: Each trainable copy of ORT is replaced by single convolution at refinement stage. 
\item \emph{V3 "directly train base model"}: The network is only pre-trained, without the refinement learning phase.
\item \emph{E1 "no task-specific condition"}: During the refinement phase, no additional conditions are added.
\item \emph{E2 "only observable latent"}: In the second phase, only observable latent embeddings are added.
\end{itemize}

The comparison results on REAL275 are summarized in Table.~\ref{table_abl-train}, by comparing the performance of these variants, it is obvious that each design plays a positive role. Particularly, comparing "V1", "V2" and "V3", the refinement training process improves all performances. The reason is that the trainable copy of pretrained backbone enables learning the correlation between priori shapes and observations. Besides, compared to the variants "E1" and "E2", the shape/observable latent embedding $f$ improves estimation accuracy. Ultimately, our complete scheme "Full" proves the proposed training strategy is effevtive for achieving the self-supervised pose estimation and shape reconstruction. 

\begin{figure*}[ht]
    \centering
    \includegraphics[width=\textwidth]{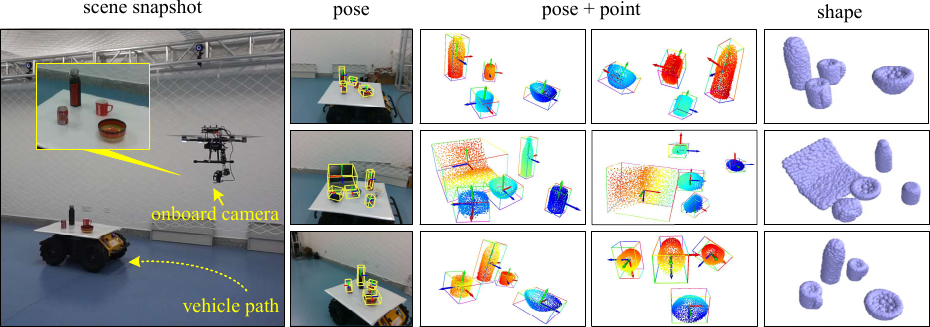}   
    \caption{\textbf{Visualisation results of aerial dynamic scene on our dataset.} Middle parts show the shape reconstruction from different viewpoints.}\label{FIG_vision-our}
\end{figure*}

\begin{table}[ht]
\scriptsize
\footnotesize
\setlength{\tabcolsep}{2.2pt}
\centering
\renewcommand\arraystretch{1.1}
\caption{\textbf{Statistical comparisons of 6-DoF pose estimation and 3D shape reconstruction on our self-built dataset.} We compare it with state-of-the-art baseline (CenterSnap~\cite{irshad2022centersnap}). The "std" is standard deviation, and the results of CenterSnap are based on its official checkpoint and code from original paper.}
\label{table_real}

\begin{tabular}{c|c|cc|cc|cc|cc}
\toprule [0.8pt]
\multirow{2}{*}{Method} & \multirow{2}{*}{Category} & \multicolumn{2}{c|}{$\text{IoU}50 \uparrow $} & \multicolumn{2}{c|}{${5^ \circ }5\,cm \uparrow $} & \multicolumn{2}{c|}{${10^ \circ }5\,cm \uparrow $} & \multicolumn{2}{c}{$\text{CD} \downarrow $} \\  
 &  & \multicolumn{1}{c}{mean} & std & \multicolumn{1}{c}{mean} & std & \multicolumn{1}{c}{mean} & std & \multicolumn{1}{c}{mean} & std \\ \midrule
\multirow{5}{*}{\begin{tabular}[c]{@{}c@{}}CenterSnap\\ \cite{irshad2022centersnap}\end{tabular}} & Bowl & \multicolumn{1}{c|}{86.2} & 3.03 & \multicolumn{1}{c|}{11.4} & 2.78 & \multicolumn{1}{c|}{28.0} & \textbf{0.21} & \multicolumn{1}{c|}{0.12} & 0.91 \\
                                         & Bottle & \multicolumn{1}{c|}{89.8} & 4.17 & \multicolumn{1}{c|}{9.8} & 3.99 & \multicolumn{1}{c|}{25.7} & \textbf{0.39} & \multicolumn{1}{c|}{0.23} & \textbf{0.56} \\
                                         & Can & \multicolumn{1}{c|}{89.7} & 2.42 & \multicolumn{1}{c|}{15.4} & 3.31 & \multicolumn{1}{c|}{33.6} & \textbf{0.57} & \multicolumn{1}{c|}{0.15} & \textbf{0.34} \\
                                         & Mug & \multicolumn{1}{c|}{90.0} & \textbf{1.47} & \multicolumn{1}{c|}{14.8} & \textbf{0.57} & \multicolumn{1}{c|}{18.6} & 0.44 & \multicolumn{1}{c|}{0.15} & \textbf{0.28} \\
                                         & Laptop & \multicolumn{1}{c|}{86.9} & \textbf{1.47} & \multicolumn{1}{c|}{16.9} & \textbf{0.57} & \multicolumn{1}{c|}{25.3} & 0.48 & \multicolumn{1}{c|}{0.27} & \textbf{0.24} \\ \midrule
\multirow{5}{*}{Ours} & Bowl & \multicolumn{1}{c|}{\textbf{99.3}} & \textbf{2.31} & \multicolumn{1}{c|}{\textbf{25.7}} & \textbf{1.63} & \multicolumn{1}{c|}{\textbf{59.9}} & 0.33 & \multicolumn{1}{c|}{\textbf{0.09}} & \textbf{0.65} \\
                      & Bottle & \multicolumn{1}{c|}{\textbf{98.3}} & \textbf{2.16} & \multicolumn{1}{c|}{\textbf{29.3}} & \textbf{2.41} & \multicolumn{1}{c|}{\textbf{71.5}} & 0.42 & \multicolumn{1}{c|}{\textbf{0.17}} & 0.79 \\
                      & Can & \multicolumn{1}{c|}{\textbf{97.5}} & \textbf{1.64} & \multicolumn{1}{c|}{\textbf{37.8}} & \textbf{1.29} & \multicolumn{1}{c|}{\textbf{72.1}} & 1.17 & \multicolumn{1}{c|}{\textbf{0.08}} & 1.77 \\
                      & Mug & \multicolumn{1}{c|}{\textbf{97.6}} & 1.65 & \multicolumn{1}{c|}{\textbf{21.3}} & 1.46 & \multicolumn{1}{c|}{\textbf{62.4}} & \textbf{0.40} & \multicolumn{1}{c|}{\textbf{0.11}} & 1.25 \\ 
                      & Laptop & \multicolumn{1}{c|}{\textbf{90.5}} & 1.88 & \multicolumn{1}{c|}{\textbf{20.1}} & 1.00 & \multicolumn{1}{c|}{\textbf{69.1}} & \textbf{0.42} & \multicolumn{1}{c|}{\textbf{0.13}} & 1.20 \\ \bottomrule [0.8pt]
\end{tabular}
\end{table}
\subsection{Comparison and Evaluation on Our Dataset}

Finally, to validate the real-world performance of our method in dynamic environments where the camera captures moving objects, we evaluate its performance on our real-world dataset with trained model. We consider five main categories: \emph{Bowl, Bottle, Can, Mug} and \emph{Laptop}.
Since these categories align with those in the REAL275 dataset, their ground-truth CAD models are also identical.
Table.~\ref{table_real} reports the performance of our method along with an available strong baseline for multi-object tasks, namely CenterSnap~\cite{irshad2022centersnap}. As it could be seen, our method shows a good performance that achieves an average accuracy of 96.6\%, 26.8\% and 67.0\% for $IoU50$, ${5^ \circ }5\,cm$ and ${10^ \circ }5\,cm$. In comparison with CenterSnap, our method improves pose estimation performance by 8.1\%, 13.1\%, and 40.8\%, respectively. Additionally, Our method achieves superior performance in 3D shape reconstruction evaluation, achieving a mAP of 0.11 for 3D shape based on the metric of CD.
Qualitative results are illustrated in Fig.~\ref{FIG_vision-our}, providing further evidence of the effectiveness of our method in practical scenarios.


\section{Limitations and future work}
Despite our method shows promising performance, it still has some limitations. For example, our model's reliance on shape priors restricts its zero-shot generalization ability to unseen instances.
In the future, we will consider two aspects: First, we plan to enhance zero-shot inference capabilities for pose tracking and shape reconstruction tasks. Second, inspired by the impressive capability of language description and large language models, we intend to investigate the potential of leveraging multimodal data to develop multimodal approaches for these tasks.

\section{Conclusion}
In this paper, we propose an effevtive self-supervised learning for multi-objective shape reconstruction and pose estimation, which only requires shape priors, eliminating the need for precisely labeling pose and CAD model for individual objects. Specifically, we introduce a diffusion-driven self-supervised network employing the Prior-Aware Pyramid 3D Point Transformer to explore the SE(3)-equivariant pose features and 3D scale-invariant shape information. 
Besides, we introduce a pretrain-to-refine self-supervised training paradigm that leverages priori shapes and shape/observation latent representation to train the proposed network. Extensive experiments conducted on four public datasets and our self-built test set show that our proposed method achieves
superior performance in both self-supervised and fully-supervised pose estimation and shape reconstruction tasks.


\ifCLASSOPTIONcompsoc
  \section*{Acknowledgments}
\else
  \section*{Acknowledgment}
\fi
This work was supported by the National Key Research and Development Program of China (No.2022YFB3903800). Jingtao Sun is supported by the China Scholarship Council under Grant No.202206130072. The work of Mike Zheng Shou is supported by the National Research Foundation, Singapore under its NRFF Award NRF-NRFF13-2021-0008, and his Start-Up Grant from National University of Singapore. Professor Ajmal Saeed Mian is the recipient of an Australian Research Council Future Fellowship Award (No.FT210100268) funded by the Australian Government.



%
\bibliographystyle{Bibliography/IEEEtran}
\bibliography{Bibliography/IEEEabrv,Bibliography/BIB_xx-TPAMI-xxxx}\ 

%

\vspace{-0.6cm}
\begin{IEEEbiography}[{\includegraphics[width=1in,height=1.25in,clip,keepaspectratio]{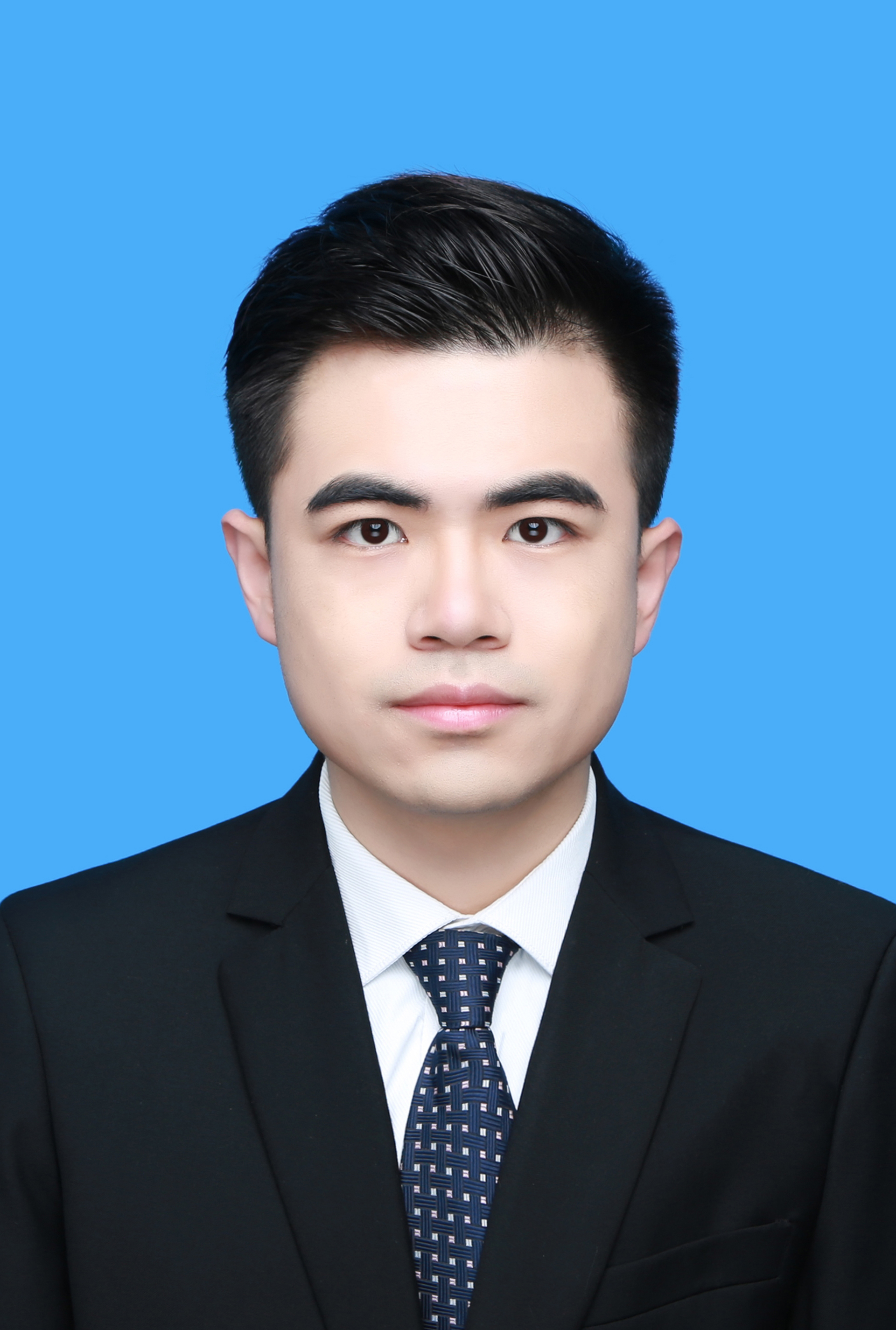}}]
{Jingtao Sun} received the B.S. and M.S. degree in College of Electrical and Information Engineering from Hunan University, Changsha, China, where he is currently working toward the Ph.D. degree with the National Engineering Research Center for Robot Visual Perception and Control, Hunan University. He is also currently a visiting Ph.D student at the Department of Electrical and Computer Engineering (ECE), National University of Singapore (NUS). His research interests include 3D computer vision, robotics and multi-modal.
\end{IEEEbiography}
\vspace{-0.6cm}
\begin{IEEEbiography}[{\includegraphics[width=1in,height=1.25in,clip,keepaspectratio]{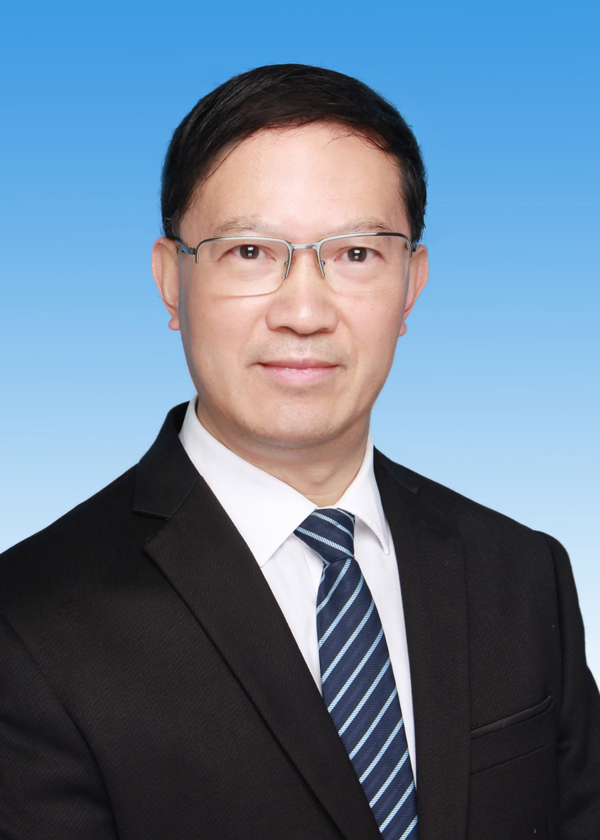}}]
{Yaonan Wang} received the Ph.D. degree in electrical engineering from Hunan University, Changsha, China, in 1994. He was a PostDoctoral Research Fellow with the Normal University of Defence Technology, Changsha, from 1994 to 1995. From 1998 to 2000, he was a Senior Humboldt Fellow in Germany, and, from 2001 to 2004, he was a Visiting Professor with the University of Bremen, Bremen, Germany. Since 1995, he has been a Professor with the College of Electrical and Information Engineering, Hunan University. He is an Academician with the Chinese Academy of Engineering. His current research interests include robotics and computer vision.
\end{IEEEbiography}
\vspace{-0.6cm}

\begin{IEEEbiography}[{\includegraphics[width=1in,height=1.25in,clip,keepaspectratio]{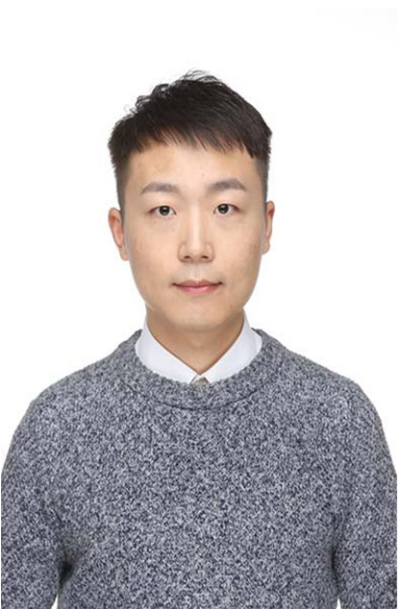}}]
{Mingtao Feng} received the Ph.D. degree at the College of Electrical and Information Engineering, Hunan University, in 2019. He was a visiting PhD student at the School of Computer Science and Software Engineering, The University of Western Australia from 2016 to 2018. He is currently an associate professor with  the School of Computer Science and Technology at Xidian University. His research interests include image processing, computer vision and machine learning.
\end{IEEEbiography}
\vspace{-0.6cm}

\vspace{-0.6cm}
\begin{IEEEbiography}[{\includegraphics[width=1in,height=1.25in,clip,keepaspectratio]{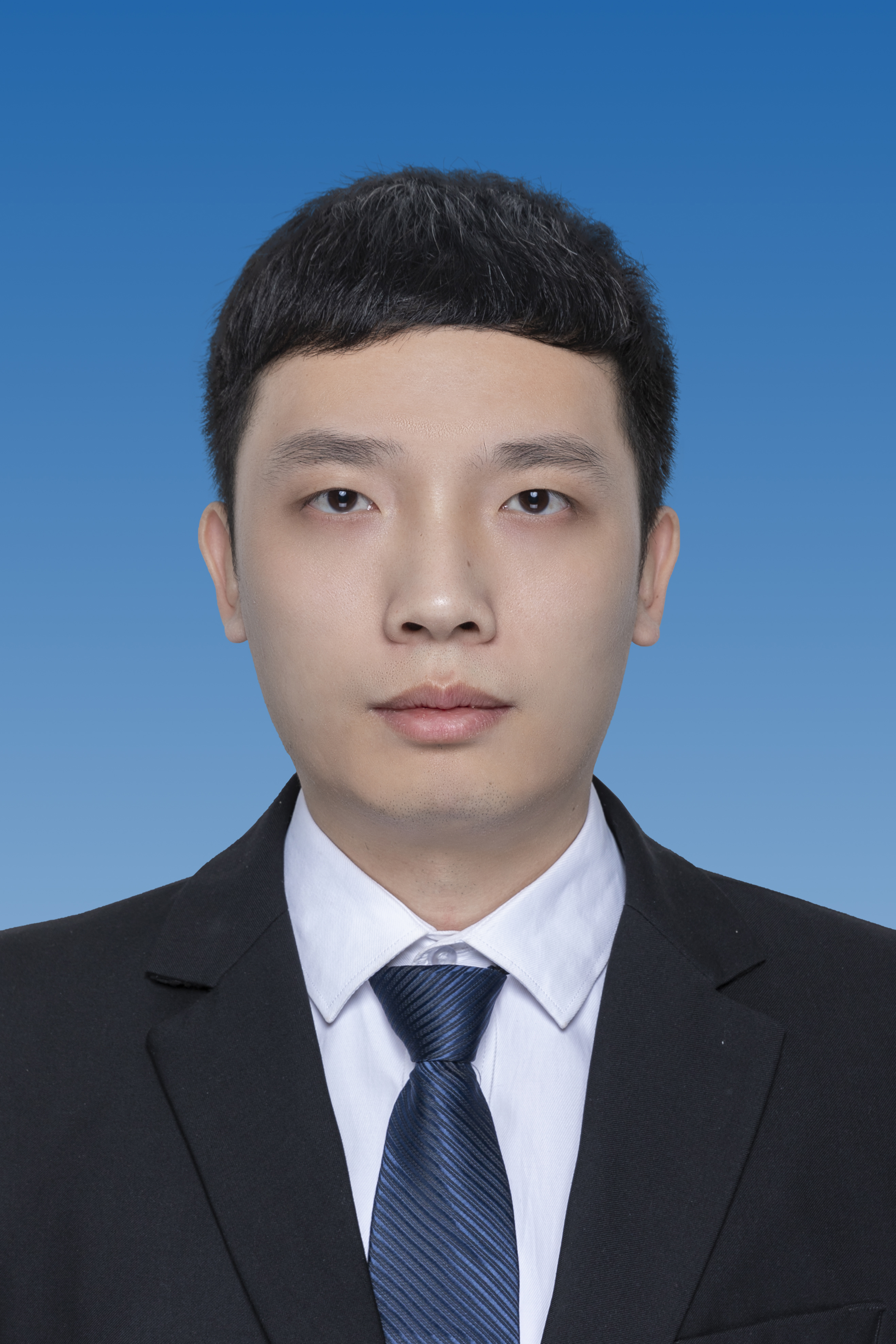}}]
{Chao Ding} received the M.S. degree in software engineering from the Central South University, Changsha, China, in 2021. He is currently pursuing a Ph.D. degree with the National Engineering Research Center for Robot Visual Perception and Control, Hunan University, Changsha, China. 
He is also currently a visiting Ph.D student at the School of Physical and Mathematical Sciences, Nanyang Technological University (NTU).
His research interests include quantum machine learning and machine learning.
\end{IEEEbiography}

\vspace{-0.6cm}
\begin{IEEEbiography}[{\includegraphics[width=1in,height=1.25in,clip,keepaspectratio]{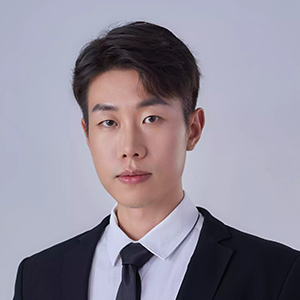}}]
{Mike Zheng Shou} reveived the Ph.D degree in Electrical Engineering from Columbia University, New York City and prior to joining NUS, he was a Research Scientist at Facebook AI, Menlo Park, California.He is currently a tenure-track Assistant Professor with an award of the National Research Foundation (NRF) Fellowship (Class of 2021) with the Department of Electrical and Computer Engineering (ECE) at National University of Singapore (NUS). He was awarded Wei Family Private Foundation Fellowship. He received the best paper finalist at CVPR'22, the best student paper nomination at CVPR'17. His team won the 1st place in the international challenges including ActivityNet 2017, Ego4D 2022, EPIC-Kitchens 2022. He is a Fellow of National Research Foundation (NRF) Singapore. He is on the Forbes 30 Under 30 Asia list. He serves as Area Chair for top-tier artificial intelligence conferences including CVPR, ECCV, ICCV, ACM MM. His research interests include computer vision and multi-modal.
\end{IEEEbiography}

\vspace{-0.6cm}
\begin{IEEEbiography}[{\includegraphics[width=1in,height=1.25in,clip,keepaspectratio]{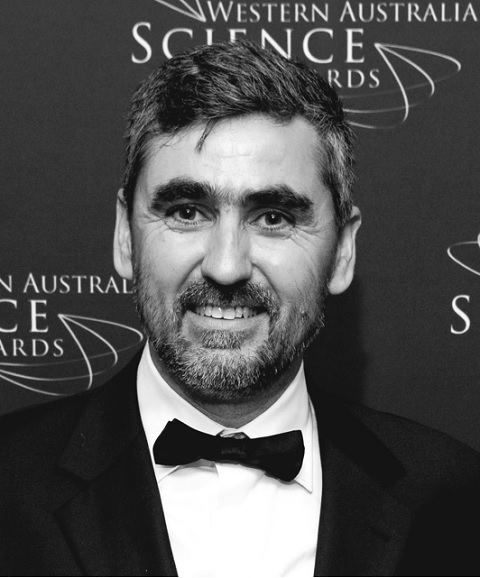}}]
{Ajmal Saeed Mian} is a Professor of Computer Science at The University of Western Australia. He has received several awards including the West Australian Early Career Scientist of the Year Award, the Aspire Professional Development Award, the Vice-chancellors Mid-career Research Award, the Outstanding Young Investigator Award, IAPR Best Scientific Paper Award, EH Thompson Award, and excellence in research supervision award. He has received several major research grants from the Australian Research Council and the National Health and Medical Research Council of Australia with a total funding of over \$13 Million. He serves as an Senior Editor of IEEE Transactions on Neural Networks and Learning Systems (TNNLS), Associate Editor of IEEE Transactions on Image Processing (TIP) and the Pattern Recognition journal (PR). He serves as Area Chair for ECCV2024, ECCV 2022, CVPR 2022, ACM MM 2024, ACM MM 2020, etc.
His research interests include computer vision, machine learning, 3D shape analysis, human action recognition, video description and hyperspectral image analysis. 
\end{IEEEbiography}

\newpage
\onecolumn
\appendices
\section{Detailed Derivations for Formulas}\label{appendix-a}

We first present the detailed content of the training objective in Eq.~(\ref{TO}), which are depicted from~Eq.~(\ref{D-TO}) to Eq.~(\ref{D-TO-1}). Next, the detailed derivations in Eq.~(\ref{D-TO-1}) are shown in Eq.~(\ref{adapt-d}). Note that we rewrite Eq.~(\ref{adapt-d}) to Eq.~(\ref{adapt-2}) due to the ${P_r}$ is fixed and is no effect on the whole Markov chain during the forward diffusion process, where each posterior q(*) is known and it is similar to the unconditional generative model:
\begin{align}\label{adapt}
 q(x_i^{(t - 1)}|x_i^{(t)},x_i^{(0)},{P_r}) = q(x_i^{(t - 1)}|x_i^{(t)},x_i^{(0)}),
\end{align}
\begin{align}\label{adapt}
q(x_i^{(T)}|x_i^{(0)},{P_r}) = q(x_i^{(T)}|x_i^{(0)}).
\end{align}

Meanwhile, considering the distribution with respect to entire point cloud ${X^{(t)}} = \{ x_i^{(t)} \in {\mathbb{R}^3}\} _{i = 1}^N$ at per timestep and the distribution of single points are independent from the overall distribution of entire point cloud ${X^{(t)}}$, we treat the distribution of point cloud as the simple product of each distribution of single points and set up ${p_\theta }(x_i^{(T)}) = {p_\theta }({X^{(T)}})$ since ${p_\theta }(x_i^{(T)})$ is a standard Gaussian distribution. Based on such analysis, we rewrite the Eq.~(\ref{adapt-2}) to be Eq.~(\ref{adapt-3}) and obtain the final training objective.
\begin{figure*}[!htp]
{\noindent} 
\begin{equation}\label{D-TO}
\begin{array}{l}
 - \log {p_\theta }(x_i^{(0)}) \le  - \log {p_\theta }(x_i^{(0)}) + {D_{KL}}(q(x_i^{(1:T)}|x_i^{(0)})||{p_\theta }(x_i^{(1:T)}|x_i^{(0)},f,{P_r}))\\[5mm]
 =  - \log {p_\theta }(x_i^{(0)}) + {\mathbb{E}_{x_i^{(1:T)} \sim q(x_i^{(1:T)}|x_i^{(0)})}}\left[ {\log \frac{{q(x_i^{(1:T)}|x_i^{(0)})}}{{{{{p_\theta }(x_i^{(0:T)}|f,{P_r})} \mathord{\left/
 {\vphantom {{{p_\theta }(x_i^{(0:T)}|f,{P_r})} {{p_\theta }(x_i^{(0)})}}} \right.
 \kern-\nulldelimiterspace} {{p_\theta }(x_i^{(0)})}}}}} \right]\\[5mm]
 =  - \log {p_\theta }(x_i^{(0)}) + {\mathbb{E}_q}\left[ {\log \frac{{q(x_i^{(1:T)}|x_i^{(0)})}}{{{p_\theta }(x_i^{(0:T)}|f,{P_r})}} + \log {p_\theta }(x_i^{(0)})} \right]\\[5mm]
 = {\mathbb{E}_q}\left[ {\log \frac{{q(x_i^{(1:T)}|x_i^{(0)})}}{{{p_\theta }(x_i^{(0:T)}|f,{P_r})}}} \right]
\end{array}
\end{equation}  
\end{figure*}
\begin{figure*}[!htp]
{\noindent} 
\begin{equation}\label{D-TO-1}
Let~~{\mathbb{E}_q}\left[ { - \log {p_\theta }(x_i^{(0)})} \right] =  - {\mathbb{E}_q}\log {p_\theta }(x_i^{(0)}) \le {\mathbb{E}_q}\left[ {\log \frac{{q(x_i^{(1:T)}|x_i^{(0)})}}{{{p_\theta }(x_i^{(0:T)}|f,{P_r})}}} \right] = L
\end{equation}  
\end{figure*} 

\begin{figure*}[!htp]
{\noindent} 
\begin{equation}\label{adapt-2}
L = {\mathbb{E}_q}\left[ {\sum\limits_{t = 2}^T {\log \frac{{q(x_i^{(t - 1)}|x_i^{(t)},x_i^{(0)})}}{{{p_\theta }(x_i^{(t - 1)}|x_i^{(t)},f,{P_r})}} + \log \frac{{q(x_i^{(T)}|x_i^{(0)})}}{{{p_\theta }(x_i^{(T)})}} - \log {p_\theta }(x_i^{(0)}|x_i^{(1)},f,{P_r})} } \right]
\end{equation}  
\end{figure*} 

\begin{figure*}[htb]
{\noindent} 
\begin{equation}\label{adapt-3}
\begin{array}{c}
L = {\mathbb{E}_q}\left[ {\sum\limits_{t = 2}^T {\log \frac{{\prod\limits_{i = 1}^N {q(x_i^{(t - 1)}|x_i^{(t)},x_i^{(0)})} }}{{\prod\limits_{i = 1}^N {{p_\theta }(x_i^{(t - 1)}|x_i^{(t)},f,{P_r})} }} + \log \frac{{\prod\limits_{i = 1}^N {q(x_i^{(T)}|x_i^{(0)})} }}{{{p_\theta }({X^{(T)}})}} - \log \prod\limits_{i = 1}^N {{p_\theta }(x_i^{(0)}|x_i^{(1)},f,{P_r})} } } \right]\\[5mm]
 = {\mathbb{E}_q}\left[ {\sum\limits_{t = 2}^T {\log \prod\limits_{i = 1}^N {\frac{{q(x_i^{(t - 1)}|x_i^{(t)},x_i^{(0)})}}{{{p_\theta }(x_i^{(t - 1)}|x_i^{(t)},f,{P_r})}}}  + \log \frac{{q({X^{(T)}}|{X^{(0)}})}}{{{p_\theta }({X^{(T)}})}} - \log \prod\limits_{i = 1}^N {{p_\theta }(x_i^{(0)}|x_i^{(1)},f,{P_r})} } } \right]\\[5mm]
 = {\mathbb{E}_q}\left[ {\sum\limits_{t = 2}^T {\sum\limits_{i = 1}^N {\underbrace {{D_{KL}}(q(x_i^{(t - 1)}|x_i^{(t)},x_i^{(0)})||{p_\theta }(x_i^{(t - 1)}|x_i^{(t)},f,{P_r}))}_{{L_{t - 1}}}} }  + \underbrace {{D_{KL}}(q({X^{(T)}}|{X^{(0)}})||{p_\theta }({X^{(T)}}))}_{{L_T}} - \sum\limits_{i = 1}^N {\underbrace {\log {p_\theta }(x_i^{(0)}|x_i^{(1)},f,{P_r})}_{{L_0}}} } \right]
\end{array}
\end{equation}  
\end{figure*}

\begin{figure*}[!ht]
{\noindent} 
\begin{equation}\label{adapt-d}
\begin{array}{l}
L = {\mathbb{E}_q}\left[ {\log \frac{{q(x_i^{(1:T)}|x_i^{(0)})}}{{{p_\theta }(x_i^{(0:T)}|f,{P_r})}}} \right]\\[5mm]
 = {\mathbb{E}_q}\left[ {\log \frac{{\prod\limits_{t = 1}^T {q(x_i^{(t)}|x_i^{(t - 1)},{P_r})} }}{{{p_\theta }(x_i^{(T)})\prod\limits_{t = 1}^T {{p_\theta }(x_i^{(t - 1)}|x_i^{(t)},f,{P_r})} }}} \right]\\[5mm]
 = {\mathbb{E}_q}\left[ {\sum\limits_{t = 1}^T {\log \frac{{q(x_i^{(t)}|x_i^{(t - 1)},{P_r})}}{{{p_\theta }(x_i^{(t - 1)}|x_i^{(t)},f,{P_r})}} - \log {p_\theta }(x_i^{(T)})} } \right]\\[5mm]
 = {\mathbb{E}_q}\left[ {\sum\limits_{t = 2}^T {\log \frac{{q(x_i^{(t)}|x_i^{(t - 1)},{P_r})}}{{{p_\theta }(x_i^{(t - 1)}|x_i^{(t)},f,{P_r})}} + \log \frac{{q(x_i^{(1)}|x_i^{(0)},{P_r})}}{{{p_\theta }(x_i^{(0)}|x_i^{(1)},f,{P_r})}} - \log {p_\theta }(x_i^{(T)})} } \right]\\[5mm]
 = {\mathbb{E}_q}\left[ {\sum\limits_{t = 2}^T {\log \left( {\frac{{q(x_i^{(t - 1)}|x_i^{(t)},x_i^{(0)},{P_r})}}{{{p_\theta }(x_i^{(t - 1)}|x_i^{(t)},f,{P_r})}} * \frac{{q(x_i^{(t)}|x_i^{(0)},{P_r})}}{{q(x_i^{(t - 1)}|x_i^{(0)},{P_r})}}} \right) + \log \frac{{q(x_i^{(1)}|x_i^{(0)},{P_r})}}{{{p_\theta }(x_i^{(0)}|x_i^{(1)},f,{P_r})}} - \log {p_\theta }(x_i^{(T)})} } \right]\\[5mm]
 = {\mathbb{E}_q}\left[ {\sum\limits_{t = 2}^T {\log \frac{{q(x_i^{(t - 1)}|x_i^{(t)},x_i^{(0)},{P_r})}}{{{p_\theta }(x_i^{(t - 1)}|x_i^{(t)},f,{P_r})}} + \sum\limits_{t = 2}^T {\log \frac{{q(x_i^{(t)}|x_i^{(0)},{P_r})}}{{q(x_i^{(t - 1)}|x_i^{(0)},{P_r})}}}  + \log \frac{{q(x_i^{(1)}|x_i^{(0)},{P_r})}}{{{p_\theta }(x_i^{(0)}|x_i^{(1)},f,{P_r})}} - \log {p_\theta }(x_i^{(T)})} } \right]\\[5mm]
 = {\mathbb{E}_q}\left[ {\sum\limits_{t = 2}^T {\log \frac{{q(x_i^{(t - 1)}|x_i^{(t)},x_i^{(0)},{P_r})}}{{{p_\theta }(x_i^{(t - 1)}|x_i^{(t)},f,{P_r})}} + \log \frac{{q(x_i^{(T)}|x_i^{(0)},{P_r})}}{{q(x_i^{(1)}|x_i^{(0)},{P_r})}} + \log \frac{{q(x_i^{(1)}|x_i^{(0)},{P_r})}}{{{p_\theta }(x_i^{(0)}|x_i^{(1)},f,{P_r})}} - \log {p_\theta }(x_i^{(T)})} } \right]\\[5mm]
 = {\mathbb{E}_q}\left[ {\sum\limits_{t = 2}^T {\log \frac{{q(x_i^{(t - 1)}|x_i^{(t)},x_i^{(0)},{P_r})}}{{{p_\theta }(x_i^{(t - 1)}|x_i^{(t)},f,{P_r})}} + \log \frac{{q(x_i^{(T)}|x_i^{(0)},{P_r})}}{{{p_\theta }(x_i^{(0)}|x_i^{(1)},f,{P_r})}} - \log {p_\theta }(x_i^{(T)})} } \right]\\[5mm]
 = {\mathbb{E}_q}\left[ {\sum\limits_{t = 2}^T {\log \frac{{q(x_i^{(t - 1)}|x_i^{(t)},x_i^{(0)},{P_r})}}{{{p_\theta }(x_i^{(t - 1)}|x_i^{(t)},f,{P_r})}} + \log \frac{{q(x_i^{(T)}|x_i^{(0)},{P_r})}}{{{p_\theta }(x_i^{(T)})}} + \log \frac{{{p_\theta }(x_i^{(T)})}}{{{p_\theta }(x_i^{(0)}|x_i^{(1)},f,{P_r})}} - \log {p_\theta }(x_i^{(T)})} } \right]\\[5mm]
 = {\mathbb{E}_q}\left[ {\sum\limits_{t = 2}^T {\log \frac{{q(x_i^{(t - 1)}|x_i^{(t)},x_i^{(0)},{P_r})}}{{{p_\theta }(x_i^{(t - 1)}|x_i^{(t)},f,{P_r})}} + \log \frac{{q(x_i^{(T)}|x_i^{(0)},{P_r})}}{{{p_\theta }(x_i^{(T)})}} - \log {p_\theta }(x_i^{(0)}|x_i^{(1)},f,{P_r})} } \right]
\end{array}
\end{equation}  
\end{figure*} 
 
\newpage



\end{document}